\documentclass[lettersize,journal]{IEEEtran}
\usepackage{amsmath,amsfonts}
\usepackage{algorithmic}
\usepackage{algorithm}
\usepackage{array}
\usepackage[caption=false,font=normalsize,labelfont=sf,textfont=sf]{subfig}
\usepackage{textcomp}
\usepackage{stfloats}
\usepackage{url}
\usepackage{verbatim}
\usepackage{graphicx}
\usepackage{cite}
\usepackage{hyperref}
\usepackage{amssymb}
\usepackage{multirow}
\usepackage{booktabs}

\usepackage{color,soul}
\sethlcolor{yellow} 
\usepackage{empheq}

\usepackage[many]{tcolorbox}
\usepackage{ctable}

\tcbset{
  myhlight/.style={
    colback=yellow,
    arc=0pt,
    outer arc=0pt,
    boxrule=0pt,
    top=2pt,
    bottom=2pt,
    left=2pt,
    right=2pt,
  },
  highlight math style={myhlight}
}

\newtcolorbox{myhl}{
  breakable,
  myhlight
}

\begin{document}

\title{Robust Proxy: Improving Adversarial Robustness by Robust Proxy Learning}

\author{Hong Joo Lee and Yong Man Ro,~\IEEEmembership{Senior Member,~IEEE,}
\thanks{This work was conducted by Center for Applied Research in Artificial Intelligence (CARAI) grant funded by DAPA and ADD (UD230017TD).}
\thanks{H. J. Lee, and Y. M. Ro are with the Image and Video Systems
Laboratory, School of Electrical Engineering, Korea Advanced Institute of
Science and Technology (KAIST), Daejeon, 34141, South Korea (e-mail:
dlghdwn008@kaist.ac.kr; ymro@kaist.ac.kr). \textit{Corresponding author: Yong Man Ro.}}
}

\markboth{Journal of \LaTeX\ Class Files,~Vol.~14, No.~8, August~2021}%
{Shell \MakeLowercase{\textit{et al.}}: A Sample Article Using IEEEtran.cls for IEEE Journals}


\maketitle

\begin{abstract}
Recently, it has been widely known that deep neural networks are highly vulnerable and easily broken by adversarial attacks. To mitigate the adversarial vulnerability, many defense algorithms have been proposed. Recently, to improve adversarial robustness, many works try to enhance feature representation by imposing more direct supervision on the discriminative feature. However, existing approaches lack an understanding of learning adversarially robust feature representation. In this paper, we propose a novel training framework called Robust Proxy Learning. In the proposed method, the model explicitly learns robust feature representations with robust proxies. To this end, firstly, we demonstrate that we can generate class-representative robust features by adding class-wise robust perturbations. Then, we use the class representative features as robust proxies. With the class-wise robust features, the model explicitly learns adversarially robust features through the proposed robust proxy learning framework. Through extensive experiments, we verify that we can manually generate robust features, and our proposed learning framework could increase the robustness of the DNNs.
\end{abstract}


\section{Introduction}
\IEEEPARstart{R}{ecently}, Deep Neural Networks (DNNs) have achieved great performance in various machine learning tasks \cite{krizhevsky2012imagenet,hannun2014deep,mikolov2013distributed}. Despite the phenomenal success of DNNs, they are highly vulnerable to adversarial attacks \cite{goodfellow2014explaining,nguyen2015deep,madry2017towards,szegedy2013intriguing,chen2017zoo,carlini2017towards}. By adding small and imperceptible perturbation to input data (\textit{i.e.,} adversarial examples), adversarial perturbations can effectively fool DNNs. Such vulnerability of DNNs could lead to security problems and the loss of the reliability of DNNs.

To mitigate the potential threat of adversarial attacks, a number of defenses have been proposed \cite{madry2017towards,liao2018defense,dhillon2018stochastic,guo2017countering,pang2019improving,lee2022masking,wang2019improving,zhang2019theoretically,rade2022reducing}. Among the existing defenses, Adversarial Training (AT) has been demonstrated to be the most effective defense strategy \cite{bai2021recent,athalye2018obfuscated}. It trains DNNs with adversarial examples by solving min-max optimization problems between the adversarial perturbation and model parameters.

However, some recent works started tackling the limitation of existing adversarial training schemes through the lens of feature representation \cite{mao2019metric,fan2021does,zhong2019adversarial,kim2020adversarial,NEURIPS2020_ba7e36c4,gowal2021selfsupervised}. They claim that well-generalized feature representation could improve the adversarial robustness. To enhance the feature representation, they apply AT framework to a self-supervised or unsupervised learning scheme such as SimCLR \cite{kim2020adversarial, NEURIPS2020_ba7e36c4}. However, they just applied the existing representation learning framework to AT framework, there is a lack of a deeper understanding of adversarially robust feature representation.

In the lens of adversarial robustness, it has been known that the disagreement between standard and adversarial robustness stems from differently trained features representation \cite{ilyas2019adversarial,kim2021distilling,roth2019odds}. The vulnerability of DNNs arises from naturally learned non-robust feature components, and they are highly correlated with adversarial prediction. Ilyas et al. \cite{ilyas2019adversarial} demonstrated that features consist of robust and non-robust components. They claimed that the adversarial examples are directly attributed to the presence of non-robust components and these non-robust components are brittle by adversarial perturbations while useful for prediction in the standard setting. Kim et al. \cite{kim2021distilling} explicitly distilled features into the robust and non-robust components. Specifically, they disentangled features into the robust and non-robust channels. Then, they showed that the vulnerability mainly stems from non-robust channels rather than robust channels. The aforementioned analyses of the adversarial vulnerability commonly argue that the feature representations that are learned to correctly predict and robustly predict are different. Therefore, to improve adversarial robustness, it is necessary to learn adversarially robust feature representations. Although there are many studies to identify the robust and non-robust features, only a few works have been conducted to exploit these robust and non-robust features.

Based on the aforementioned adversarially robust feature representation view, in this paper, we raise the following intriguing, yet thus far overlooked questions: 
\begin{center}
{``\textit{How can we make DNNs learn the adversarially robust feature?}"}
\end{center} To address the question, in this paper, we generate class representative robust features for all classes. Then, with the generated features, we train DNNs to explicitly learn adversarially robust features. 

To generate the robust feature, firstly, we employ a feature distillation method proposed in \cite{kim2021distilling} and distill the feature into robust and non-robust channels. With the distilled features, we quantify the effect of the non-robust channels on the prediction by the gradient of them. Then, we optimize the input to minimize the magnitude of the gradient of non-robust channels. Specifically, we add a Robust Perturbation (RP, $r$) to input data and optimize the perturbation to reduce the gradient of non-robust channels. After that, we extend the process of optimizing robust perturbation to the process of optimizing Class-wise Robust Perturbation (CRP, $r^k$). The CRP is added to any input data corresponding to the target class $k$ and makes the input robust against adversarial perturbation. To optimize CRP, we exploit the Empirical Risk Minimization (ERM) optimization which is considered a successful recipe for finding classifiers with small population risk. Specifically, we propose a novel optimization process called Class-wise ERM Optimization (CEO). In the CEO algorithm, we quantify the empirical risk of the target class as the expectation of gradient for the non-robust channels. Then, we optimize CRP to reduce the gradient. Through empirical and theoretical analysis, we show that CRP makes the corresponding class input not easily attacked by adversarial perturbation and provides a robust prediction. 

With the optimized CRP, we generate a class-representative robust feature called Robust Proxy. To generate a robust proxy, we randomly sample each class image from the training dataset and add CRP to the corresponding class images. Then, we extract features from the CRP-added images and use the features as robust proxies. During the training, DNNs explicitly learn the representation of robust proxy through the proposed robust proxy learning framework. For each proxy, we pull the data of the same class close to the proxy and push others away in the feature space, allowing the model to explicitly learn adversarially robust features.

The major contributions of the paper are as follows.

\begin{itemize}
    \item We propose a novel way to generate adversarially robust features by optimizing robust perturbation. Then, we extend the robust perturbation to class-wise robust perturbation to generate class-representative robust features.
    \item With the CRP, we train DNNs with the proposed learning framework called Robust Proxy Learning. In the proposed method, we train the DNNs to explicitly learn adversarially robust features by using robust proxies.
    \item Through extensive experiments, we show that we could explicitly learn robust features and improve the robustness.
\end{itemize}

\section{Related Work}
\subsection{Understanding Adversarially Robust Features}
As adversarial vulnerability has attracted significant attention, many works are devoted to getting to the bottom of the vulnerability \cite{ilyas2019adversarial,kim2021distilling,roth2019odds}. An early study tended to view adversarial examples as a result of the excessive linearity nature of DNNs in high-dimensional spaces \cite{goodfellow2014explaining}. In another study, it has been regarded as statistical fluctuations in the data manifold \cite{szegedy2013intriguing,shafahi2018are}.

Recently, a new perspective on the phenomenon of adversarial vulnerability is proposed \cite{roth2019odds,tsipras2018robustness,ilyas2019adversarial,kim2021distilling}. In contrast to previous studies, these works figure out the vulnerability as a view of feature representations. Tsipras et al. \cite{tsipras2018robustness} showed that the goals of learning features for standard accuracy and adversarial robustness might be at odds. Specifically, they argue that the features for adversarial robustness and for standard accuracy are fundamentally different. Then, the features learned by robust models tend to align better with salient data characteristics and human perception. Ilyas et al. \cite{ilyas2019adversarial} demonstrated that adversarial vulnerability is a consequence of non-robust components of features. These non-robust components are useful and highly predictive for standard performance, yet easily broken by adversarial perturbations. In contrast, robust components still can provide robust prediction results even with adversarial perturbation. Kim et al. \cite{kim2021distilling} explicitly distilled features into robust channels and non-robust channels. Then, they showed that non-robust channels are directly related to adversarial predictions. Also, in \cite{kim2021distilling}, they addressed that the robust channels are robust on the noise variation and invariant to the existence of the adversarial perturbation. In contrast, the non-robust channels are brittle and easily change the model prediction by noise variation. In the rest of the paper, we use the definition of robust/non-robust features defined in \cite{kim2021distilling}.

The aforementioned studies commonly argued that features for a standard performance and for adversarial robustness are different. Also, the vulnerability stems from the non-robust feature components that are brittle and incomprehensible to humans.  Therefore, it is necessary to learn robust feature representations. In this context, recently some works try to improve the robustness by exploiting the robust feature representations. Yang et al. \mbox{\cite{yang2021adversarial}} proposed the Deep Robust Representation Disentanglement Network (DRRDN) model to disentangle the class-specific representation and class-irrelevant representation. To this end, they employed a disentangler to extract and align the robust representations from both adversarial and natural examples. With the disentangler, they eliminate the effect of adversarial perturbations and improve the robustness. Kim et al. \mbox{\cite{Kim_2023_CVPR}} proposed a way to extract robust and non-robust features based on causality. They demystified causal features on adversarial examples in order to uncover inexplicable adversarial origins through a causal perspective. To this end, they proposed adversarial instrumental variable (IV) regression as a means to identify the causal features pertaining to the causal relationship of adversarial prediction on adversarial examples. Then, they improve the robustness by exploiting the causal features.

In this work, we propose a new approach that explicitly learns robust feature representations rather than heuristically employing existing learning algorithms.

\subsection{Adversarial Training}
Adversarial Training (AT) is one of the most effective approaches to defending against adversarial attacks \cite{madry2017towards,wang2019improving,zhang2019theoretically,rade2022reducing,zhang2020geometry,jakubovitz2018improving}. By solving a min-max optimization between model parameters and adversarial perturbation, it improves the adversarial robustness. Madry et al. \cite{madry2017towards} proposed a PGD-based adversarial training method and achieved the first empirical adversarial robustness. Since then, it became a milestone in adversarial training methods.

\textbf{AT by Enhancing Feature Representation:} Many recent studies tackled the limitation of existing adversarial training methods in the lens of feature representation \cite{mao2019metric,fan2021does,zhong2019adversarial,kim2020adversarial,NEURIPS2020_ba7e36c4,gowal2021selfsupervised}. Then, they started to study how to improve feature representation by exploiting self-supervised / unsupervised learning scheme. They tried to apply AT framework to a self-supervised / unsupervised pretraining task to make DNNs learn robust data representation. Mao et al. \cite{mao2019metric} empirically analyze the feature representations under adversarial attack and showed that adversarial perturbations shift the feature representations of adversarial examples away from their true class and closer to the false class. With the empirical observations, they employ triplet-wise distance loss in the AT framework and improve the robustness. Fan et al. \cite{fan2021does} proposed a unified adversarial contrastive learning framework that learns well transferable feature representations. Similar works have been done to enhance the feature representation against adversarial attacks \cite{Zhong_2019_ICCV,zhong2019adversarial,kim2020adversarial,wang2021agkd}.

However, since they heuristically employ existing representation learning framework to AT framework, they do not take any consideration of the adversarially robust feature representations into account. It has been known that naturally learned feature representation that aims to correctly predict is different from adversarially robust feature representation. Therefore, it is necessary to consider the adversarially robust feature representation to improve the adversarial robustness.

\section{Generating Class Representative Robust Features}
The main contribution of this paper is to generate class-representative robust features. Then, we train the model robustly by using the class representative robust features. In this section, we first describe how to generate class-representative robust features. Specifically, to clarify how to distill robust and non-robust channels from features, we briefly revisit \cite{kim2021distilling}. Then, with the distilled features, we explain how to generate adversarially robust features and class representative robust features. Through the proof concept experiments, we verify that we could generate class-representative robust features by using Class-wise Robust Perturbations (CRP).

\subsection{Revisiting Robust \& non-Robust Feature Distillation}
In this section, we revisit how to distill features into robust and non-robust channels. In this paper, the definition of robust and non-robust channels stems from \cite{kim2021distilling}. Let $z$ indicates the intermediate feature of the model $f$ such that $z=f_{l}(x)$, where $f_l(\cdot)$ is $l$-th layer output of the given model. Then, $f_{l+}(\cdot)$ represents subsequent network after the $l$-th layer. Therefore, the prediction of the model can be written as $y’=f_{l+}(f_{l}(x))$. Since the last convolution layer contains higher-level features and more discriminative features than the earlier convolution layer, we identify the robust features at the last convolutional layer before the global average pooling layer or fully-connected layer.
Here, the feature $z$ has $C$ channels and each channel has inherent feature variation. The $\sigma_{z}$ indicates inherent feature variation of the feature $z$ for each channel. Therefore, it can be written as $\sigma_{z}=[\sigma_{z,1},\sigma_{z,2},…, \sigma_{z,C}]$ (given parameters). With the given parameter ($\sigma_{z}$), we set the criterion for comparison as $T=max(\sigma^{2}_{z})$, where $T$ denotes the maximum tolerance of the noise variation of the original feature. Here, we use $T$ as the threshold that discriminates robust and non-robust channels. Then, we find the noise variation $\sigma$ to estimate the prediction sensitivity of each feature channel along the noise intervention. If the noise variation $\sigma^2_c>T$, the channel is regarded as a robust channel. To find the noise variation ($\sigma$), in \cite{kim2021distilling}, they exploited information bottleneck \cite{alemi2017deep}. According to the definition of information bottleneck in \mbox{\cite{kim2021distilling, alemi2017deep}}, it could find maximally informative representation for target labels, restraining input information, concurrently. Therefore, by using the information bottleneck, we can quantify the feature importance and information flow for the target labels. Then, we can quantify the importance of each feature by utilizing the information bottleneck. The optimization objective can be written as follow:

\begin{equation}
    \min_{\sigma}L=\underbrace{-y\cdot log(f_{l+}(f_{l}(x)+\sigma \cdot \epsilon))}_{\text{cross-entropy}}+\underbrace{\beta D_{KL}[p(z|x)|q_{\sigma}(z))]}_{\text{KL divergence}}.
    \label{eq:1}
\end{equation}

The first term indicates cross-entropy and the second term indicates KL divergence between the original feature and noise added feature. Also, the $\epsilon$ indicates Gaussian noise, and $y$ denotes the ground-truth label. Through the optimization process, we find noise variation $\sigma=[\sigma_1, \sigma_2,…, \sigma_C]$ (optimized parameters). In \cite{kim2021distilling}, they have analyzed that $\sigma$ is related to variance in Gaussian, thus large $\sigma$ can allow for large variation capacity in the channel, which makes it have the ability to overcome feature variation. On the other way, small $\sigma$ only allows for a small variation capacity in the channel, which makes it brittle to feature variation. 

After we optimize $\sigma$, following the aforementioned criterion, we find the robust channel index ($i_r=[i_{r,1},i_{r,2},…, i_{r,C}]$). If the optimized noise variation $\sigma^2_c>T$, the channel is regarded as robust channel and $i_{r,c}=1$. Then, the non-robust channels are simply reversed from the robust channel index such that $i_{nr,c}=1-i_{r,c}$. Finally, the set of robust channel features can be written as $z_{r}=z \cdot i_{r}$. Similarly, the set of non-robust channel feature $z_{nr}$ can be written as $z_{nr}=z \cdot i_{nr}$. In this way, the $z$ can be expressed as $z=i_{r}\cdot z+i_{nr}\cdot z$.


\subsection{Generating Robust Features}
\subsubsection{Problem Definition} In this section, we describe how to generate robust features. To generate robust features, we generate robust inputs and use the features extracted from the inputs as robust features. To this end, we manipulate the input by adding perturbation and define the perturbation as Robust Perturbation (RP, $r$), which makes the input robust against adversarial perturbation. Then, the input is regarded as the robust input if the model prediction maintains its original prediction, even though there exists adversarial perturbation. Therefore, the problem can be defined as follows,

\begin{equation}
\begin{aligned}
    \text{Making} \quad &f(x+r+\delta^*) \approx f(x) \text{ by optimizing } r\\
    \text{where} \quad &\delta^*=\underset{||\delta||<\epsilon}{\textrm{argmax}}\mathcal{L}(f(x+r+\delta)),y),
\end{aligned}
\label{eq:2}
\end{equation}

\noindent where $\delta^*$ is an adversarial perturbation that attacks $x+r$, $\epsilon$ is adversarial perturbation budget, $x$ is an input image, and $\mathcal{L}$ denotes the objective function such as cross-entropy. The equation can be interpreted that even though the input $x+r$ is attacked by the adversarial perturbation, the prediction is maintained and the feature extracted from $x+r$ can be regarded as a robust feature. In the following section, we will explain how to optimize $r$ in detail.

\begin{figure}[!t]
	\centering
	    \includegraphics[width=0.98\linewidth]{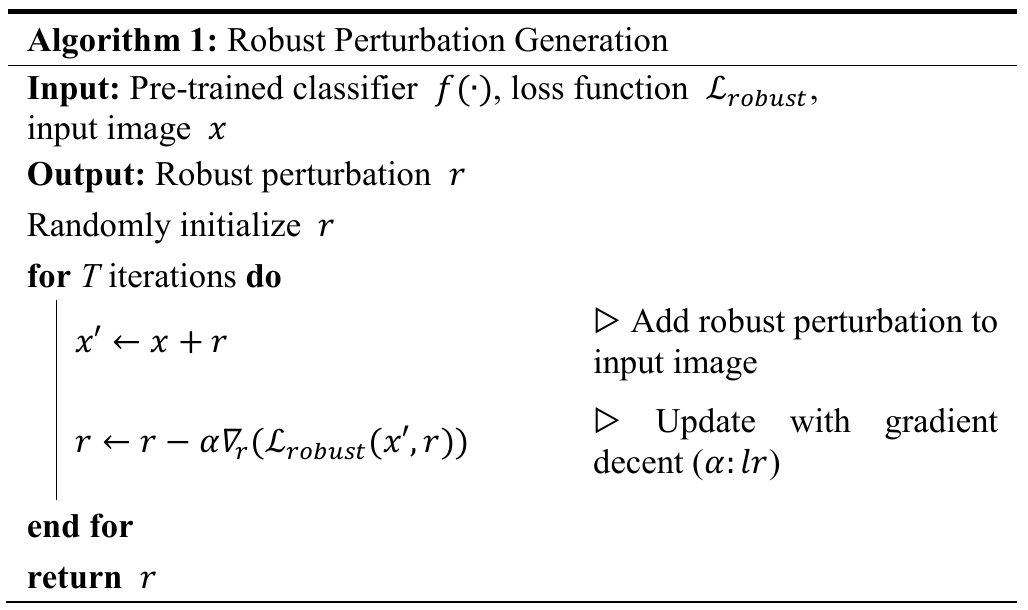}
\label{alg:1}
\end{figure}

\subsubsection{Optimizing Robust Perturbation}
Recall that the adversarial vulnerability mainly stems from non-robust components of feature \cite{ilyas2019adversarial,tsipras2018robustness,kim2021distilling}. Specifically, in \cite{kim2021distilling}, the vulnerability is highly correlated with the non-robust channel of $z$. If the non-robust channels have a large impact on the prediction, it can be interpreted that the input is vulnerable to adversarial perturbation. Therefore, it is necessary to reduce the influence of non-robust channels on prediction. To this end, in this paper, we quantify the influence of non-robust channels by measuring the gradient of the cost function with respect to non-robust channels and reducing them by optimizing $r$. Note that, the gradient of the non-robust channels represents how small changes at each channel affect the prediction. The gradient of non-robust features can be described as follows:

\begin{equation}
\mathcal{G}_{nr}=\frac{\partial}{\partial z_{nr}}\mathcal{L}_{base}(f(x+r),y),
\label{eq:3}
\end{equation}
where $z_{nr}$ is the set of non-robust channels in feature $z=\{z_{r}\cup z_{nr}\}$, $f(\cdot)$ is a model prediction, and $\mathcal{L}_{base}$ is a loss function to ensure correct prediction. We define the loss function as $\mathcal{L}_{base}=-c\cdot \max(\underset{i\neq y}{\max}(f(x)_i)-f(x)_i,0)$ due to its empirical effectiveness of optimization performance in \cite{madry2017towards}. $c$ determines the trade-off between the size of the perturbation added to the input and the degree of correct prediction in Eq.\mbox{\ref{eq:5}}. $\mathcal{G}_{nr}$ is a quantified value that quantifies how much a change in the non-robust feature changes the correct prediction. If $\mathcal{G}_{nr}$ has a large value, it indicates that we can easily change the prediction for the ground-truth class. The gradient of non-robust channel can be simplified as follows:

\begin{equation}
\begin{aligned}
\mathcal{G}_{nr}&=\frac{\partial}{\partial z_{nr}}\mathcal{L}_{base}(f(x+r),y)\\
&=\frac{\partial z}{\partial z_{nr}}\frac{\partial}{\partial z}\mathcal{L}_{base}(f(x+r),y)\\
&=i_{nr}\cdot\frac{\partial}{\partial z}\mathcal{L}_{base}(f(x+r),y).\\
&\quad\quad\quad(\because z=i_{r}\cdot z+i_{nr}\cdot z)
\end{aligned}
\label{eq:4}
\end{equation}

Following Eq.\ref{eq:4}, we could simply reduce the gradient of non-robust channels by multiplying non-robust channel index ($i_{nr}$) to the gradient of objective function respect to the feature ($\frac{\partial}{\partial z}\mathcal{L}_{base}(\cdot)$). Using the gradient, we optimize robust perturbation by optimizing the following objective:
\begin{equation}
\mathcal{L}_{robust}=\mathcal{L}_{base}(f(x+r),y)+\left\|\mathcal{G}_{nr} \right\|_{2} + \left\|r \right\|_{2}.
\label{eq:5}
\end{equation}
Optimizing Eq. \ref{eq:5} means that we find a small and imperceptible perturbation that makes the model predict well and reduces the gradient of non-robust features. Then, reducing the gradients of non-robust channels contains the same effect of reducing $\sigma_{nr}$ to resist adversarial perturbation. Therefore, features extracted from those inputs ($x+r$) can be a robust feature. Note that there is no regularization parameter before the gradient norm of non-robust features and the norm of perturbation in Eq.\mbox{\ref{eq:5}} The optimization algorithm is described in Algorithm 1.


\subsubsection{Robustness Analysis with Robust Perturbation}
The goal of optimizing $r$ is to make the input itself robust against adversarial perturbation. If we can robustify the input by adding $r$ to the input, we can extract features from these inputs and regard them as robust features. For verification, in this section, we conduct a proof concept experiment to verify whether we can make the input itself robust against adversarial attacks by augmenting the input.

In the proof concept experiment, we optimize $r_i$ for all corresponding images $x_i$ in the test dataset of CIFAR-10 and Tiny-ImageNet on the pre-trained AT models (Madry \cite{madry2017towards}, TRADES \cite{zhang2019theoretically}, MART \cite{wang2019improving}, and HELP \cite{rade2022reducing}). The results are described in Table \ref{tab:1}. Table \ref{tab:1} shows the robustness comparison under the PGD-20 attack according to different input types. In the table, $x_i$ denotes an accuracy when the adversarial perturbation is added to the original input image, and $x_i+r_i$ denotes an accuracy when the adversarial perturbation is added to $x_i+r_i$. Note that to verify that $r_i$ truly makes the input robust, we generate the adversarial perturbation on $x_i+r_i$. As shown in the table, we verify that adding $r_i$ to input data can significantly improve the robustness. Furthermore, even though the adversarial perturbation is generated on $x_i+r_i$, $r_i$ could successfully improve the robustness (robust against adaptive attack). This can be interpreted that $r_i$ does not cause gradient obfuscation \cite{carlini2017towards}, and reducing the gradient of the non-robust channels makes the input robust itself.

\begin{table}[!t]
\centering
\caption{Robustness comparison when robust perturbation $r$ is added to input data. Note that the adversarial perturbation is generated by PGD-20 on $x_i+r_i$.}
\begin{tabular}{cccc}
\specialrule{.15em}{.1em}{.1em}
 Model                       & Input Types & CIFAR-10 & Tiny-ImagNet \\ \cmidrule{1-4}
\multirow{2}{*}{Madry}     & $x_i$           & 46.5       & 20.2           \\
                        & $x_i+r_i$         & \textbf{75.3}       & \textbf{48.7}           \\ \cmidrule{1-4}
\multirow{2}{*}{TRADES} & $x_i$           & 48.8       & 21.3           \\
                        & $x_i+r_i$         & \textbf{77.91}       & \textbf{46.5}            \\ \cmidrule{1-4}
\multirow{2}{*}{MART}   & $x_i$           & 49.1       & 21.2           \\
                        & $x_i+r_i$         & \textbf{79.8}       & \textbf{49.2}           \\ \cmidrule{1-4}
\multirow{2}{*}{HELP}   & $x_i$           & 52.1       & 21.6           \\
                        & $x_i+r_i$         & \textbf{78.5}       & \textbf{48.4}           \\ \specialrule{.15em}{.1em}{.1em}
\end{tabular}%
\label{tab:1}
\end{table}

\subsection{Optimizing Class-wise Robust Perturbation}

\begin{figure*}[!t]
	\centering
	    \includegraphics[width=0.85\linewidth]{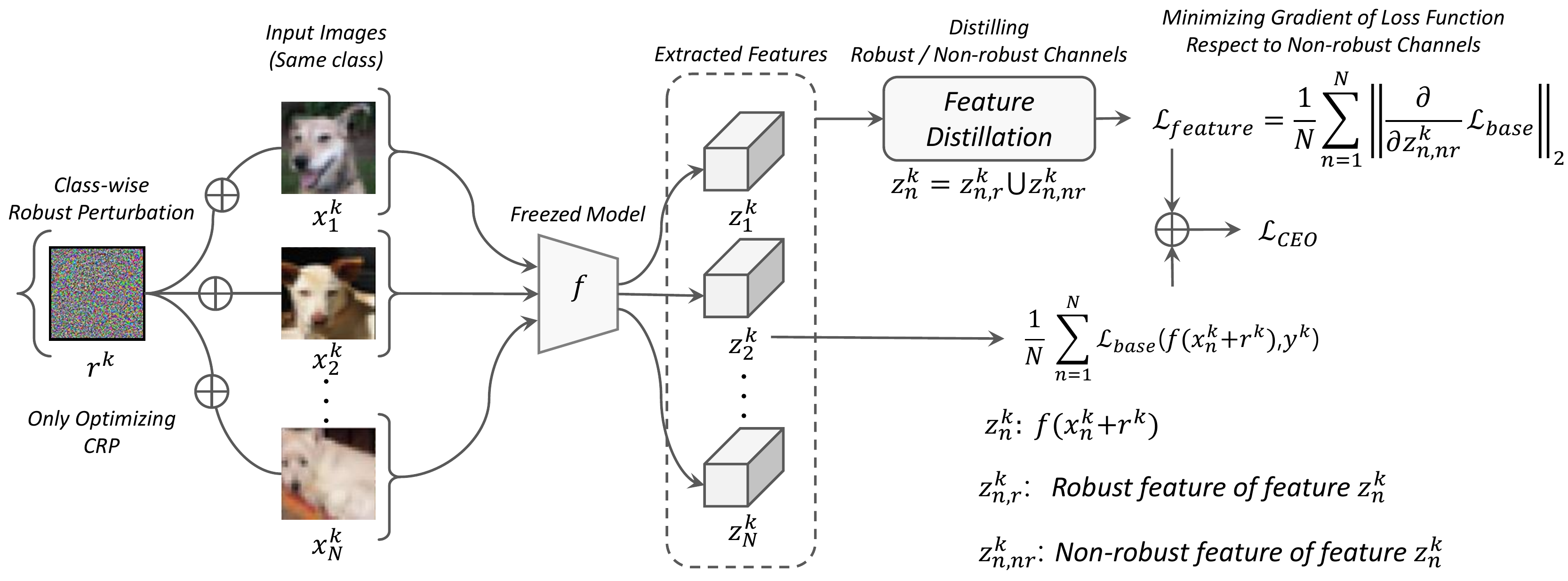}
	\caption{Overview of the proposed CEO process. The CRP is added to input for the corresponding class, fed into the freezed model, and extract feature. Then, the features are distilled into robust / non-robust channels and we measure the empirical risk of non-robust channels to optimize the CRP. The calculated objective function is backpropagated to CRP and we only optimize CRP.}
	\label{fig:2}
	\vspace{-0.3cm}
\end{figure*}

We have analyzed that augmenting input with robust perturbation makes the input robust against adversarial perturbation. In this section, we expand the concept of $r$ to Class-wise Robust Perturbation (CRP, $r^k$) where $k$ denotes the class index. The CRP is a class-specific perturbation that can be applied to any corresponding class input ($x^k$) and improve the robustness. For example, if we generate a CRP for the dog class ($r^{dog}$), it can be applied to any dog class images and improve the robustness. To this end, we propose a novel optimization method called Class-wise ERM Optimization (CEO) by extending Empirical Risk Minimization (ERM) algorithm, which is considered a successful recipe for finding classifiers with small population risk \cite{madry2017towards}. In order to make $r^k$ have universality in corresponding class inputs, we measure the empirical risk of non-robust channels for corresponding class images. Let $X=\{x^k_{1},x^k_{2},...,x^k_{N}\}$ be a subset of class-$k$ images sampled from the training data and the feature extracted from each input is $z^{k}_{n}$. Then, the empirical risk of non-robust channels can be formulated as follows:
\begin{equation}
\mathcal{L}_{feature}= \frac{1}{N}\sum_{n=1}^{N}\left\|\frac{\partial}{\partial z^{k}_{n,nr}}\mathcal{L}_{base}(x^{k}_{n}+r^{k},y^{k}_{n}) \right\|,
\label{eq:6}
\end{equation}
where $N$ denotes the number of images in the target class and $z^{k}_{n,nr}$ denotes the set of non-robust channels of $n$-th image. Therefore, the total objective function for CEO is
\begin{equation}
\mathcal{L}_{CEO}=\frac{1}{N}\sum_{n=1}^{N}\mathcal{L}_{base}(x^{k}_{n}+r^{k},y^{k}_{n}) +\mathcal{L}_{feature}.
\label{eq:7}
\end{equation}
The objective function means that the CRP is optimized to correctly classify the target class $k$ and reduce the effect of the non-robust channels within the class. Here, the gradient of CEO with non-robust channels is gradually converged to local optima, where it asymptotically closes to zero vector to make it be small population risk. Through the CEO, we could generate CRP that could improve the robustness of the target class and generate class representative common robust features. Fig. \ref{fig:2} gives an overview of how to optimize the CRP.

\begin{table}[!t]
\centering
\caption{Robustness comparison when class-wise robust perturbation $r^k$ is added to input data. Note that the adversarial perturbation is generated  by PGD-20 on $x^{k}_{i}+r^k$.}
\begin{tabular}{cccc}
\specialrule{.15em}{.1em}{.1em}
   Model                       & Input Types & CIFAR-10 & Tiny-ImagNet \\ \cmidrule{1-4}
\multirow{2}{*}{AT}     & $x^{k}_{i}$           & 46.5       & 20.2           \\
                        & $x^{k}_{i}+r^{k}$         & \textbf{69.2}       & \textbf{35.3}           \\ \cmidrule{1-4}
\multirow{2}{*}{TRADES} & $x^{k}_{i}$           & 48.8       & 21.3           \\
                        & $x^{k}_{i}+r^{k}$         & \textbf{69.8}       & \textbf{35.2}            \\ \cmidrule{1-4}
\multirow{2}{*}{MART}   & $x^{k}_{i}$           & 49.1       & 21.2           \\
                        & $x^{k}_{i}+r^{k}$         & \textbf{68.27}       & \textbf{37.0}           \\ \cmidrule{1-4}
\multirow{2}{*}{HELP}   & $x^{k}_{i}$           & 52.1       & 21.6           \\
                        & $x^{k}_{i}+r^{k}$         & \textbf{70.22}       & \textbf{36.8}           \\ \specialrule{.15em}{.1em}{.1em}
\end{tabular}%
\vspace{-0.2cm}
\label{tab:2}
\end{table}

\begin{figure}[!t]
  \centering  
  \includegraphics[width=0.6\linewidth]{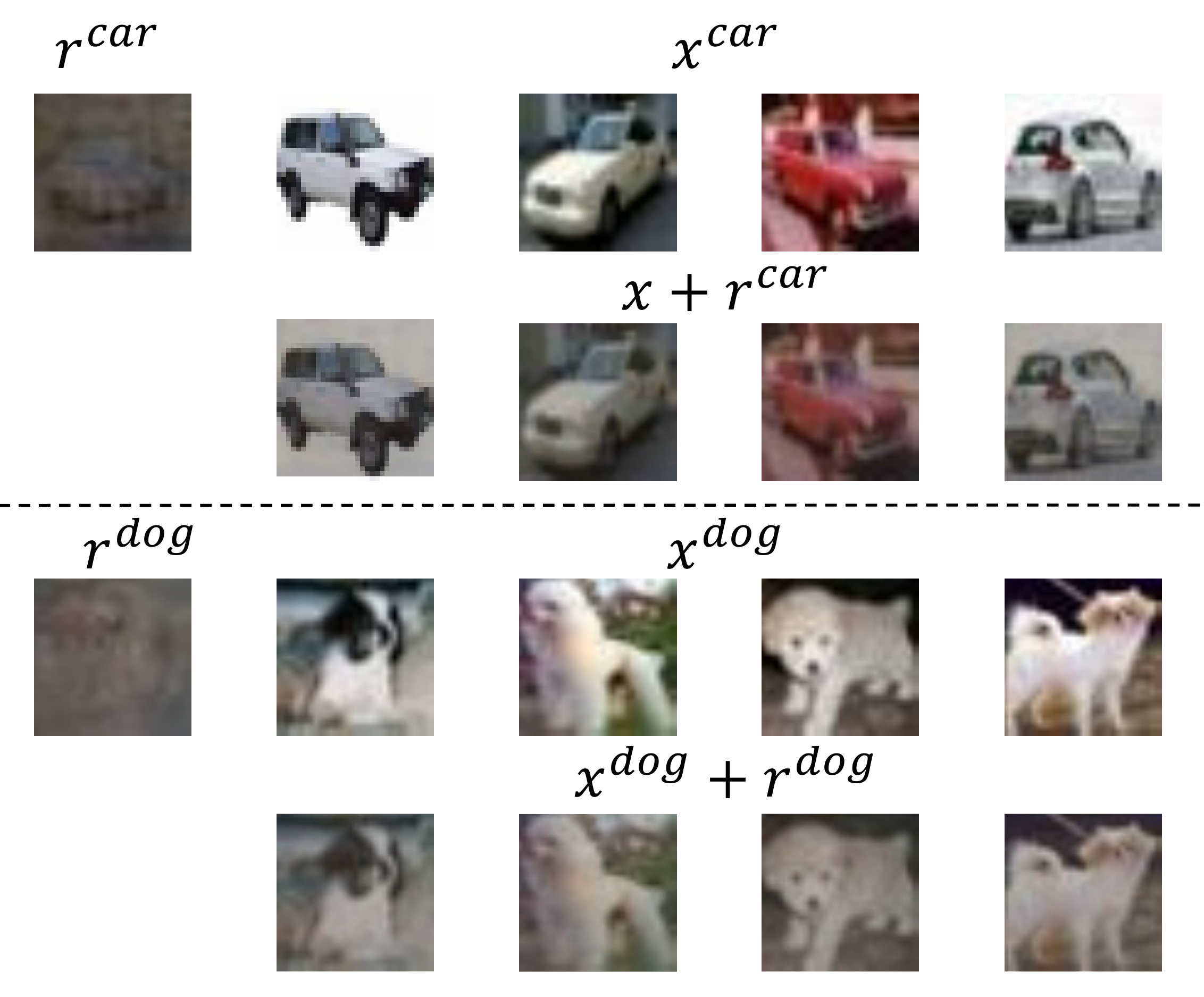}
  \vspace{-0.2cm}
  \caption{Visualization results of CRPs (\mbox{$r^k$}) and CRPs added images (\mbox{$x^k+r^k$}). \mbox{k} denotes car and dog classes.}
  \label{fig:minor_1}
\end{figure}


\begin{figure}[!t]
	\centering
	    \includegraphics[width=0.85\linewidth]{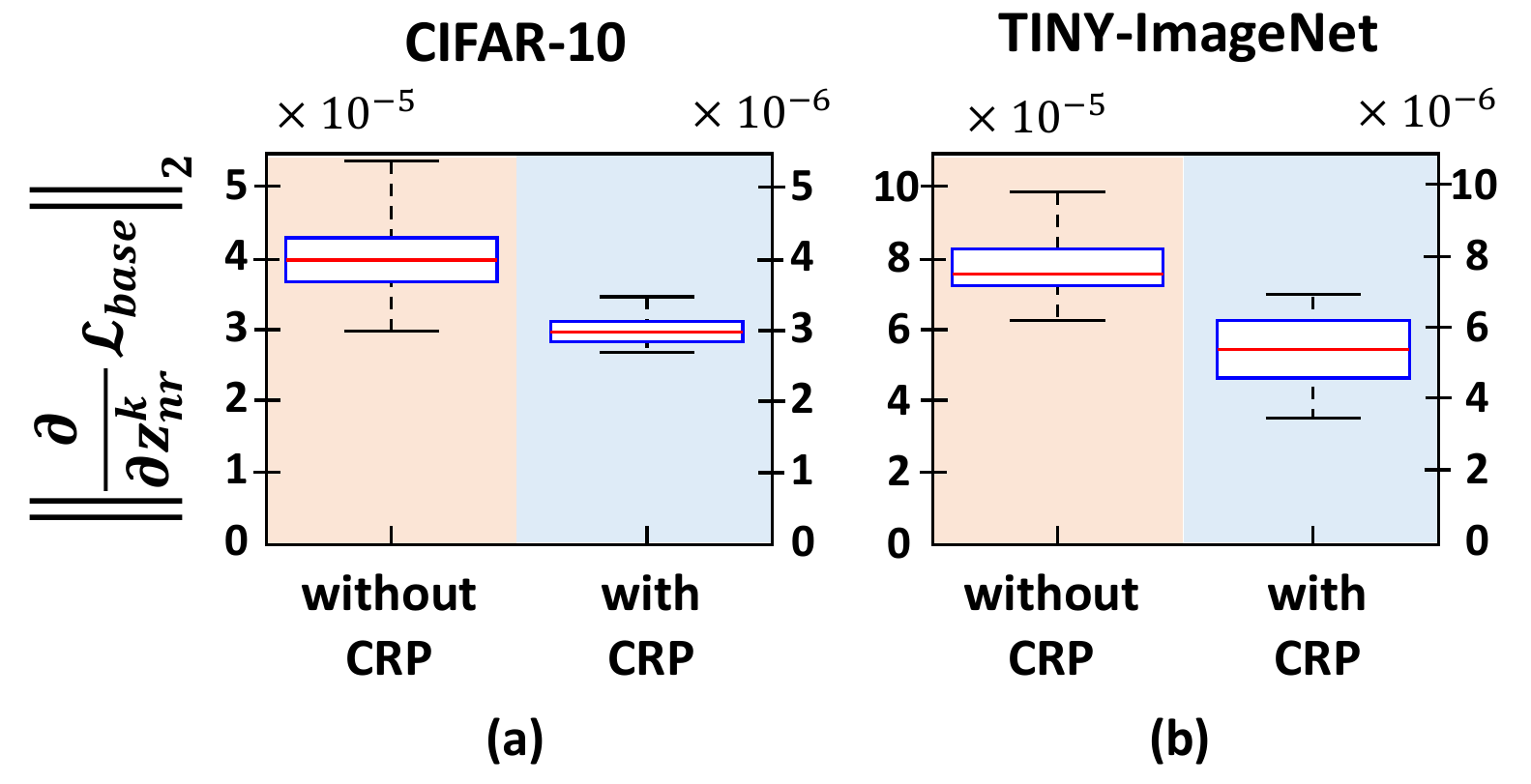}
\caption{Distribution of L2-norm of non-robust features. (a) and (b) is the non-robust feature gradient of CIFAR-10 and Tiny-ImageNet respectively.}
   	\label{fig:6}

\end{figure}

\subsubsection{Analysis of CRP}
The goal of optimizing $r^k$ is to generate the class-wise robust perturbation that could improve the robustness of the corresponding target class images. For verification, we also conduct proof concept experiments. In the experiments, we optimize $r^k$ for each class from the training set and apply them to the test set images. Table \ref{tab:2} shows the robustness comparison under the PGD-20 attack according to different input types. In the table, $x^k_i$ denotes an accuracy when we use the original input image, and $x^k_i+r^k$ denotes an accuracy when adding CRP to corresponding target class images. Different from the result in Table \ref{tab:1} that generates robust perturbation for all corresponding input images, in this experiment, we generate one CRP per class and applied it to all images that correspond class. The results are shown in Table \ref{tab:2}. As shown in the table, we verify that adding CRP to input also significantly improves the robustness. In other words, once we optimize CRP from training images, we can apply CRP to any test image that corresponds to the target class and robustify the target class images. Furthermore, we visualize the CRPs.Fig. \mbox{\ref{fig:minor_1}} shows the examples of CRPs (\mbox{$r^k$}) and CRPs added images (\mbox{$x^k+r^k$}). As shown in the figure, CRP contains the semantic information of the corresponding class, and its magnitude is very small. Therefore, even if we add the CRPs to corresponding class images, we can robustify the corresponding image well without any significant change.

\begin{figure}[!t]
  \centering  
  \includegraphics[width=0.55\linewidth]{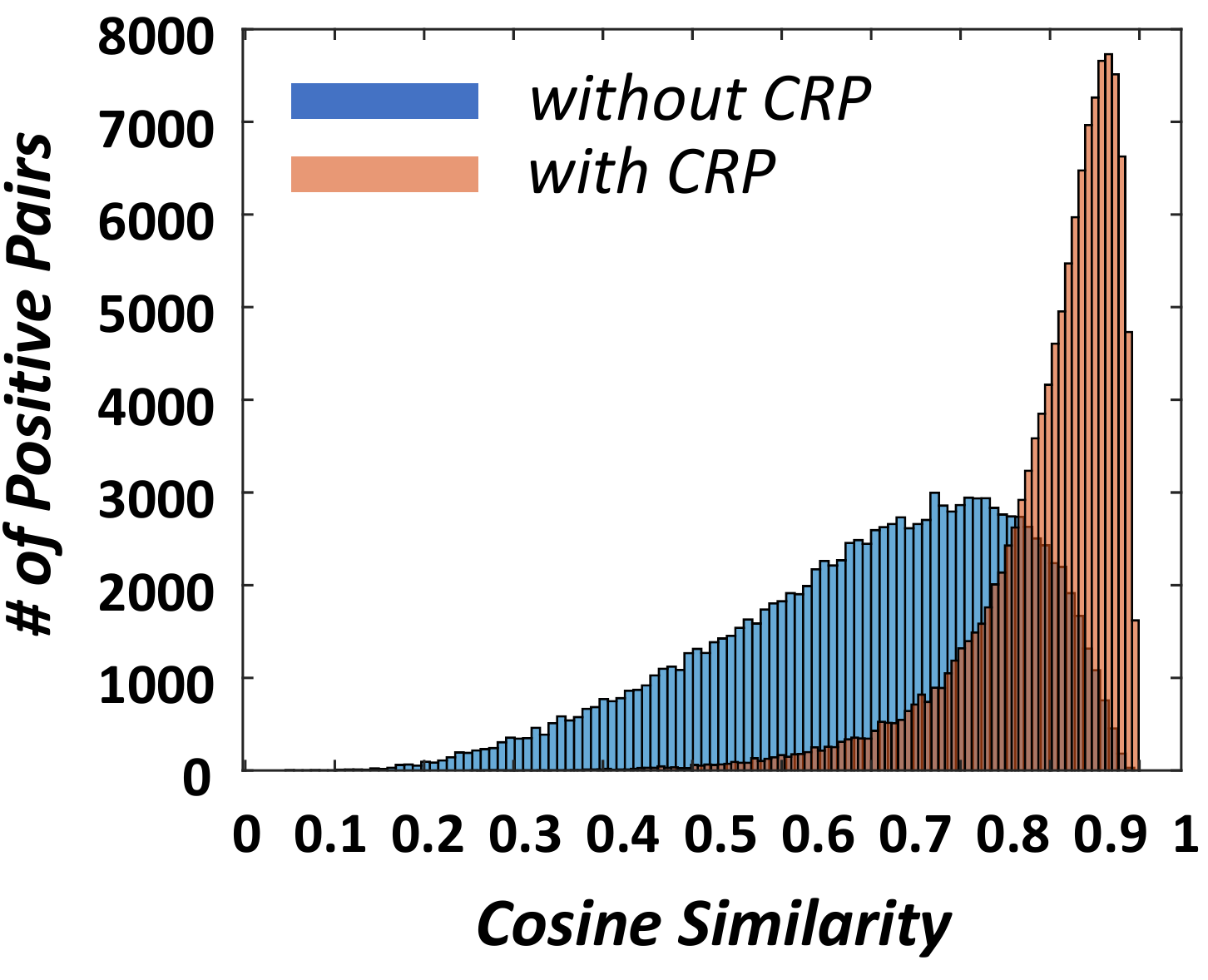}
  \caption{Feature similarity distribution between positive samples. The blue bar denotes the feature similarity distribution between images belonging to the same class. The orange bars denote the feature similarity distribution between the same classes when CRP is added to the image.}

  \label{fig:4}
\end{figure}

\textbf{\textit{Measuring the Gradient of Non-robust Channels:}}
As we discussed above, the norm of the gradients with respect to the non-robust channels are related to adversarial vulnerability. Then, we have shown that adding CRP to input can improve the adversarial robustness. In this section, we verify whether the increase in robustness is caused by the decrease in the gradient of the non-robust channels. To this end, we apply CRPs to input data and measure the L2-norm magnitude of non-robust channels. The results are shown in Fig. \ref{fig:6}. As shown in the figure, the gradients of non-robust channels are reduced significantly compared with `without CRP'. Therefore, we demonstrate that CRPs could improve the robustness by reducing the gradient of non-robust features.

\textbf{\textit{Feature Similarity Measurement:}} Furthermore, to verify whether adding CRP can generate a feature that can represent the class or not, we measure the feature similarity between samples belonging to the same class (positive samples). The result is shown in Fig. \ref{fig:4}. As shown in the figure, when applying the CRP to input data (orange color bar) the similarity between positive samples is increased and has narrow distribution. The experimental result shows that if the features are extracted by adding CRP to any input corresponding to the target class, the features have similar representations and can be regarded as robust and class-representative features.

\section{Proposed Robust Proxy Learning}
\begin{figure}[!t]
	\centering
	    \includegraphics[width=0.95\linewidth]{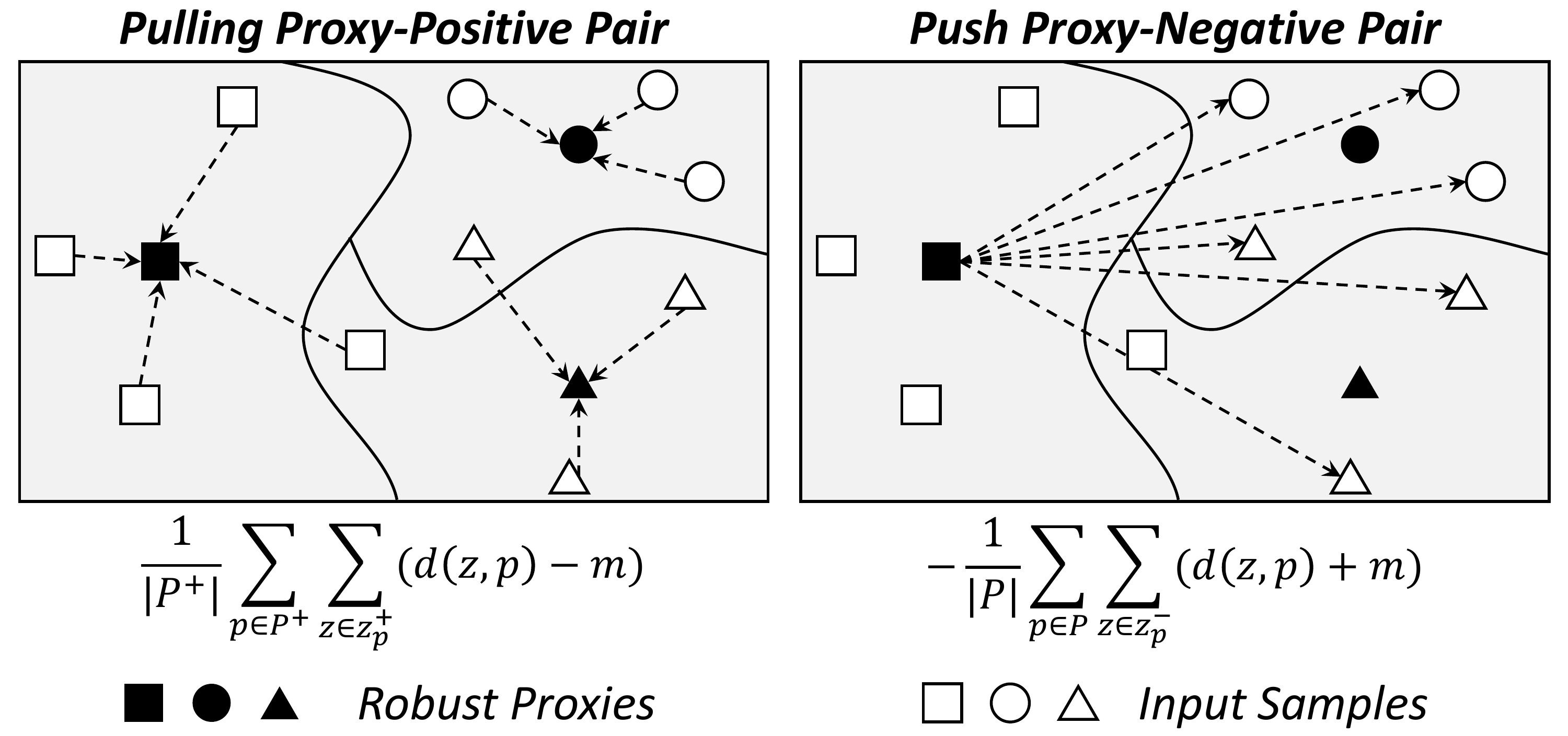}

	\caption{Graphical explanation of the proposed robust proxy learning framework. Each robust proxy is colored in black and different shapes denotes different classes. Note that for simplicity, in the case of proxy-negative pair, we only describe for one proxy.} 

	\label{fig:5}

\end{figure}
\subsection{Generating Robust Proxy}
In this section, we introduce our novel robust proxy learning framework.  The main idea is to regard each robust proxy as an anchor in Triplet loss and associate it with entire data in a batch. To this end, we first explain how to generate robust proxies. In the previous section, we verified that even if features are extracted by adding $r^k$ to any input belonging to the target class $k$, the features have similar values. Therefore, we randomly select the input corresponding to each class, add the corresponding CRP, and then extract the features. The extracted features can be regarded as class-representative robust features and we use them as robust proxies.

\begin{figure}[!t]
	\centering
	    \includegraphics[width=0.95\linewidth]{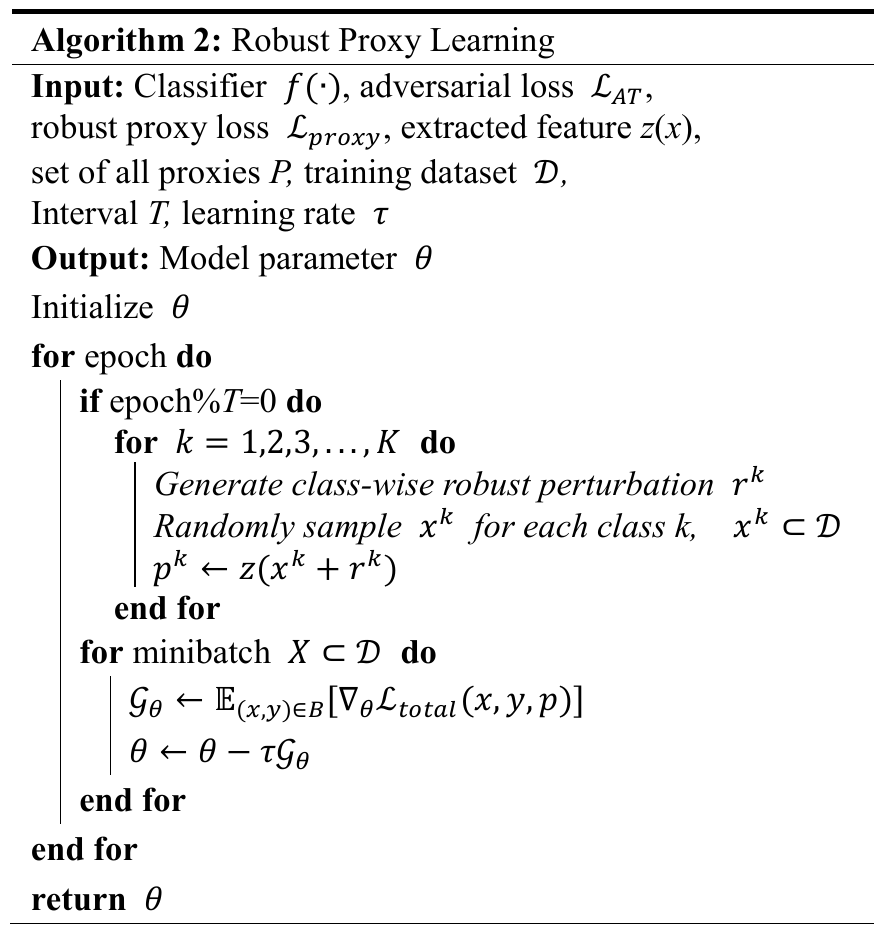}
\label{alg:2}
\vspace{-0.5cm}
\end{figure}

\begin{table*}[t]
\centering
\caption{Adversarial robustness comparison on CIFAR-10 and Tiny-ImageNet dataset under white-box attack setting. The backbone models are ResNet-18 and Wide-ResNet28-10 (WRN28-10).}
\resizebox{0.95\linewidth}{!}{
\begin{tabular}{clcccccccccc}
\specialrule{.15em}{.1em}{.1em}
\multirow{2}{*}{Backbone}  & \multirow{2}{*}{Method} & \multicolumn{5}{c}{CIFAR-10}              & \multicolumn{5}{c}{Tiny-ImageNet}         \\ \cmidrule(lr){3-7}\cmidrule(lr){8-12} 
                           &                    &Clean     & FGSM     & PGD      & CW       & AA     &Clean  & FGSM     & PGD      & CW       & AA       \\ \cmidrule{1-12}
\multirow{9}{*}{ResNet-18} & Standard   &92.1$\pm$0.10 &18.1$\pm$0.13 &1.1$\pm$0.02 &0.0$\pm$0.00 &0.0$\pm$0.00 &60.3$\pm$0.02 &6.2$\pm$0.10 &0.8$\pm$0.01 &0.0$\pm$0.00 &0.0$\pm$0.00 \\ \cmidrule{2-12}
                           & Madry                  &83.8$\pm$0.02    & 60.3$\pm$0.02     & 50.1$\pm$0.04     & 48.3$\pm$0.03     & 47.0$\pm$0.03     & 46.1$\pm$0.02   & 23.3$\pm$0.05     & 20.2$\pm$0.06     & 18.2$\pm$0.04     & 17.6$\pm$0.05     \\
                           & Madry+Proxy            &\textbf{84.5$\pm$0.02}    & \textbf{62.0$\pm$0.04}       & \textbf{53.1$\pm$0.06}       & \textbf{49.4$\pm$0.04}       & \textbf{48.9$\pm$0.04}       &\textbf{46.3$\pm$0.04}   & \textbf{24.1$\pm$0.02}       & \textbf{22.7$\pm$0.03}       & \textbf{20.5$\pm$0.03}       & \textbf{19.9$\pm$0.05}       \\ \cmidrule{2-12} 
                           & TRADES             &83.0$\pm$0.02     & 61.4$\pm$0.05     & 53.0$\pm$0.06     & 48.6$\pm$0.05     & 47.0$\pm$0.05     &\textbf{47.8$\pm$0.06}   & 24.4$\pm$0.07     & 21.3$\pm$0.02     & 18.4$\pm$0.04     & 18.2$\pm$0.06     \\
                           & TRADES+Proxy       &\textbf{84.0$\pm$0.04}     & \textbf{63.1$\pm$0.05}       & \textbf{55.3$\pm$0.04}       & \textbf{49.7$\pm$0.05}       & \textbf{49.8$\pm$0.05}       & 47.5$\pm$0.07   & \textbf{25.1$\pm$0.05}       & \textbf{22.5$\pm$0.06}       & \textbf{21.2$\pm$0.06}       & \textbf{20.0$\pm$0.05}       \\ \cmidrule{2-12} 
                           & MART               &82.3$\pm$0.06     & 60.6$\pm$0.05     & 54.1$\pm$0.08     & 48.5$\pm$0.06     & 47.2$\pm$0.06     &47.6$\pm$0.07   & 23.9$\pm$0.02     & 21.2$\pm$0.04     & 19.2$\pm$0.03     & 18.1$\pm$0.03     \\
                           & MART+Proxy         &\textbf{84.0$\pm$0.05}     & \textbf{63.4$\pm$0.04}       & \textbf{56.3$\pm$0.06}       & \textbf{51.0$\pm$0.05}       & \textbf{49.9$\pm$0.05}       &\textbf{47.8$\pm$0.04}   & \textbf{25.5$\pm$0.06}       & \textbf{23.1$\pm$0.05}       & \textbf{21.3$\pm$0.04}       & \textbf{20.0$\pm$0.02}       \\ \cmidrule{2-12} 
                           & HELP               &84.0$\pm$0.04     & 61.9$\pm$0.03     & 52.0$\pm$0.07     & 49.8$\pm$0.05     & 48.6$\pm$0.06     &48.0$\pm$0.07   & 24.0$\pm$0.05     & 21.6$\pm$0.03     & 19.3$\pm$0.03     & 17.7$\pm$0.05     \\
                           & HELP+Proxy         &\textbf{85.0$\pm$0.06}     & \textbf{63.8$\pm$0.06}       & \textbf{55.3$\pm$0.04}       & \textbf{52.0$\pm$0.04}       & \textbf{51.2$\pm$0.04}       &\textbf{49.0$\pm$0.08}   & \textbf{25.3$\pm$0.02}       & \textbf{23.0$\pm$0.05}       & \textbf{20.5$\pm$0.01}       & \textbf{19.1$\pm$0.01}       \\ \cmidrule{1-12}
\multirow{9}{*}{WRN28-10}  & Standard   &96.2$\pm$0.12 &22.3$\pm$0.10 &3.5$\pm$0.01 &0.0$\pm$0.00 &0.0$\pm$0.00 &63.1$\pm$0.02 &9.2$\pm$0.02 &2.7$\pm$0.00 &0.0$\pm$0.00 &0.0$\pm$0.00 \\ \cmidrule{2-12}
                           & Madry                 &85.5$\pm$0.02     & 62.3$\pm$0.04     & 54.2$\pm$0.05     & 50.8$\pm$0.06     & 49.9$\pm$0.04     &48.6$\pm$0.03   & 25.1$\pm$0.08     & 23.0$\pm$0.06     & 20.0$\pm$0.06     & 18.7$\pm$0.06     \\
                           & Madry+Proxy           &\textbf{86.8$\pm$0.02}     & \textbf{64.5$\pm$0.05}       & \textbf{57.8$\pm$0.05}       & \textbf{52.3$\pm$0.06}       & \textbf{51.7$\pm$0.06}       &\textbf{50.0$\pm$0.06}   & \textbf{27.2$\pm$0.04}       & \textbf{25.9$\pm$0.05}       & \textbf{22.3$\pm$0.05}       & \textbf{20.1$\pm$0.04}       \\ \cmidrule{2-12} 
                           & TRADES             &85.2$\pm$0.04     & 63.4$\pm$0.04     & 56.2$\pm$0.06     & 50.7$\pm$0.04     & 49.8$\pm$0.05     &50.6$\pm$0.07   & 26.8$\pm$0.08     & 25.1$\pm$0.02     & 21.9$\pm$0.03     & 19.0$\pm$0.01     \\
                           & TRADES+Proxy       &\textbf{86.8$\pm$0.05}     & \textbf{66.2$\pm$0.06}       & \textbf{58.3$\pm$0.03}       & \textbf{53.8$\pm$0.02}       & \textbf{52.0$\pm$0.02}       &\textbf{52.0$\pm$0.03}   & \textbf{28.3$\pm$0.02}       & \textbf{27.7$\pm$0.03}       & \textbf{22.8$\pm$0.02}       & \textbf{21.3$\pm$0.03}       \\ \cmidrule{2-12} 
                           & MART               &85.7$\pm$0.09     & 63.5$\pm$0.05     & 56.2$\pm$0.06     & 52.4$\pm$0.05     & 51.0$\pm$0.05     &50.4$\pm$0.04   & 28.2$\pm$0.04     & 26.2$\pm$0.06     & 23.4$\pm$0.03     & 20.4$\pm$0.03     \\
                           & MART+Proxy         &\textbf{86.7$\pm$0.08}     & \textbf{65.6$\pm$0.06}       & \textbf{59.4$\pm$0.04}       & \textbf{54.0$\pm$0.04}       & \textbf{53.1$\pm$0.03}       &\textbf{52.5$\pm$0.05}   & \textbf{30.8$\pm$0.05}       & \textbf{28.7$\pm$0.04}       & \textbf{24.0$\pm$0.02}       & \textbf{22.6$\pm$0.02}       \\ \cmidrule{2-12} 
                           & HELP               &86.8$\pm$0.08     & 65.0$\pm$0.06     & 56.8$\pm$0.06     & 53.9$\pm$0.05     & 52.5$\pm$0.06     &51.6$\pm$0.05   & 27.8$\pm$0.04     & 26.7$\pm$0.06     & 23.5$\pm$0.03     & 20.0$\pm$0.03     \\
                           & HELP+Proxy         &\textbf{86.0$\pm$0.05}     & \textbf{67.2$\pm$0.04} & \textbf{57.7$\pm$0.04} & \textbf{55.3$\pm$0.04} & \textbf{54.0$\pm$0.03} & \textbf{53.1$\pm$0.06} & \textbf{29.6$\pm$0.05} & \textbf{27.9$\pm$0.05} & \textbf{24.2$\pm$0.04} & \textbf{22.8$\pm$0.04} \\ \specialrule{.15em}{.1em}{.1em}
\end{tabular}
}
\vspace{-0.5cm}
\label{tab:3}
\end{table*}

\subsection{Training with Robust Proxies}

With the robust proxies, we make the DNNs explicitly learn adversarially robust features. Fig. \ref{fig:5} briefly illustrates how to learn robust features during the training. In the figure, each proxy is colored in black, and different shapes indicate different classes. As shown in the figure, positive samples and corresponding positive proxy ($p^+$) are trained to reduce the distance from each other, whereas negative samples and each proxy are trained to increase the distance from each other. This can be formulated as follows:
\begin{equation}
\begin{aligned}
    \mathcal{L}_{proxy}=&\underbrace{\frac{1}{\left|P^+ \right|}\sum_{p\in P^+}^{}\sum_{z\in z^{+}_{p}}^{}(d(z,p)-m)}_{\textrm{Pulling to the Proxy}}-\\
    &\underbrace{\frac{1}{\left|P \right|}\sum_{p\in P}^{}\sum_{z\in z^{-}_{p}}^{}(d(z,p)+m)}_{\textrm{Pushing from the Proxy}}
    \label{eq:prox}
\end{aligned}
\end{equation}
where $m$ denotes a margin, $P$ denotes the set of all proxies, $P^+$ denotes the set of positive proxies of data in a batch, and $d(\cdot)$ denotes a distance function. Also, $z^{+}_p$ denotes the set of positive features of corresponding $p$, $z^{-}_p$ denotes the set of negative features of $p^+$, and we use cosine similarity between two vectors as a distance function. To boost the robustness of existing AT methods, we jointly optimize the existing AT losses and proxy loss. Therefore, we optimize $\mathcal{L}_{total}=\mathcal{L}_{AT}+\mathcal{L}_{proxy}$ to train the model. $\mathcal{L}_{AT}$ denotes the existing loss function of AT methods. For example, in the case of MART \cite{wang2019improving} the $\mathcal{L}_{AT}$ can be written as $\mathcal{L}_{AT}=BCE(f(x+\delta),y)+\lambda(1-f_{y}(x))KL(f(x+\delta),f(x))$, where $f_y(x)$ denotes the output probability of ground-truth class, $\lambda$ is a tunable scaling parameter that balances the two parts of the loss, and BCE($\cdot$) adds the cross-entropy loss and margin loss terms to improve the decision margin of the classifier. Then, $\mathcal{L}_{proxy}$ makes the model learn features in which the effects of non-robust features are suppressed during training and separate the different classes. After training, we do not need robust perturbation for inference. Therefore, during the inference, robust perturbations are not added to the test data.

\noindent\textbf{Training Details:} In this section, we describe the training details of the proposed robust proxy learning. Note that, since robust/non-robust features can be distilled from the adversarially trained model, we initialize the model parameters of the AT model. Then, we fine-tuned the model parameter with $\mathcal{L}_{total}$ and set the margin value as 1.0. Since generating robust proxies for every epoch increase the training time, we refresh the proxies for every $T=5$ epoch. The details algorithm is described in Algorithm 2.

\section{Experiments}
\subsection{Experiment Setting}
\subsubsection{Dataset and Network} We conduct experiments to verify the effectiveness of the proposed method on two datasets (CIFAR-10 \cite{krizhevsky2009learning} and Tiny-ImageNet \cite{le2015tiny} datasets). The CIFAR-10 dataset consists of 50,000 training images and 10,000 test images with 10 classes. The Tiny-ImageNet dataset has 200 classes and each class has 500 training images, 50 validation images, and 50 test images. For both datasets, we use the ResNet-18 \cite{he2016deep} and WideResNet28-10 \cite{zagoruyko2016wide} networks. 

\subsubsection{Attack Settings} To evaluate the defensive performance of the proposed method, we conduct four adversarial attack methods used as benchmarks for evaluating adversarial robustness (FGSM \cite{goodfellow2014explaining}, PGD \cite{madry2017towards}, CW \cite{carlini2017towards}, and AutoAttack \cite{croce2020reliable}). For all attack methods, we set the perturbation budget $\epsilon=8/255$. We generate adversarial perturbation with 20 iterations with the step size $\epsilon/10$ for the PGD attack. In the case of CW \cite{carlini2017towards} attack, we use $L_\infty$-norm bounded attacks with 200 iterations. In the case of Auto Attack (AA) \cite{croce2020reliable}, it includes four attack methods (APGD-CE, APGD-DLR \cite{croce2020reliable}, FAB \cite{croce2020minimally}, and Square Attack \cite{andriushchenko2020square}) and generates adversarial perturbation by ensembling them. For FAB attack hyper-parameters, we optimize the perturbation with 100 iterations and 5 random restarts. In the case of Square attack, we fed 5000 queries for the black-box attack.


\subsection{Robustness Evaluation}
\subsubsection{White-box Evaluation}
To evaluate the effectiveness of the proposed method, we apply our proposed robust proxy learning to existing AT methods. Table \ref{tab:3} shows the robustness evaluation results under various white-box attack settings on two datasets. In the table, `Clean' denotes the results when testing with a clean dataset (test without perturbation), and `Standard' denotes the results when testing on the clean model. Note that the clean dataset denotes the original test image that any perturbations such as adversarial perturbations or class-wise robust perturbations are not added. Furthermore, the clean model denotes the model trained only with the clean dataset. In other words, the clean model is trained without adversarial examples and only uses original training data for training. As shown in the table, the clean model (Standard) shows better performance on the clean dataset compared with the adversarially trained model. However, in the case of adversarial robustness (FGSM, PGD, CW, and AA), it cannot defend against adversarial attacks. 

In terms of the adversarially trained model, the baseline results are Madry \cite{madry2017towards}, MART \cite{wang2019improving}\footnote[1]{\href{https://github.com/YisenWang/MART}{https://github.com/YisenWang/MART}}, TRADES \cite{zhang2019theoretically}\footnote[2]{\href{https://github.com/yaodongyu/TRADES}{https://github.com/yaodongyu/TRADES}}, and HELP \cite{rade2022reducing}\footnote[3]{\href{https://github.com/imrahulr/hat}{https://github.com/imrahulr/hat}}. Then, the results of our proposed proxy learning are AT+Proxy, TRADES+Proxy, MART+Proxy, and HAT+Proxy. We report the mean and variance from 5 checkpoints. As shown in the table, in the case of Madry, when trained with the proposed Robust Proxy Learning, the robustness is improved to 3.5, 3.1, 1.8, and 2.9 in FGSM, PGD, CW, and AutoAttack, respectively, when trained with the CIFAR-10 dataset on ResNet-18. Also, when we combined proxy learning with other AT methods, we can get similar results. It can be interpreted that the proposed robust proxy learning framework can be well adapted to existing AT frameworks and could improve the robustness. Furthermore, the reason why the proposed method can improve robustness is that it explicitly learns robust feature presentation while learning discriminative features between classes through robust proxy learning.

In terms of clean accuracy, the proposed method does not sacrifice clean accuracy. According to \mbox{\cite{ilyas2019adversarial}}, the robust features are useful and highly related to the target class. Therefore, exploiting the robust features does not hurt clean accuracy. Furthermore, through the proposed proxy loss (Eq. \mbox{\ref{eq:prox}}), the model learns class-discriminative representations. With the proxy loss, we pull the data of the same class close to the proxy and push others away in the feature space, allowing the model to explicitly learn class discriminative features. Therefore, the proposed method shows better performance even on clean samples.

\begin{table}[!t]
\centering
\caption{Black-box attack evaluation on CIFAR-10 dataset. The perturbation is generated on WideResNet-34-10.}

\resizebox{0.98\linewidth}{!}{
\begin{tabular}{cccccc}
\specialrule{.15em}{.1em}{.1em}
\multicolumn{2}{c}{\textbf{Method}}                         & \textbf{FGSM}  & \textbf{PGD-20} & \textbf{CW}    & \textbf{AutoAttack} \\ \hline
\multicolumn{1}{c}{\multirow{2}{*}{Madry}}  & Base & 81.37          & 82.41           & 83.02          & 82.41               \\
\multicolumn{1}{c}{}                                 & Ours & \textbf{82.12} & \textbf{83.89}  & \textbf{84.21} & \textbf{83.91}      \\ \hline
\multicolumn{1}{c}{\multirow{2}{*}{TRADES}}  & Base & 82.29          & 83.01           & 83.16          & 82.98               \\
\multicolumn{1}{c}{}                                 & Ours & \textbf{83.21} & \textbf{84.39}  & \textbf{84.84} & \textbf{84.72}      \\ \hline
\multicolumn{1}{c}{\multirow{2}{*}{MART}}   & Base & 81.76          & 82.56           & 83.09          & 82.59               \\
\multicolumn{1}{c}{}                                 & Ours & \textbf{83.37} & \textbf{84.00}  & \textbf{84.27} & \textbf{84.21}      \\ \hline
\multicolumn{1}{c}{\multirow{2}{*}{HELP}} & Base & 82.33          & 83.03           & 83.62          & 83.81               \\
\multicolumn{1}{c}{}                                 & Ours & \textbf{84.16} & \textbf{85.60}  & \textbf{85.42} & \textbf{85.88}      \\ 
\specialrule{.15em}{.1em}{.1em}
\end{tabular}}

\label{tab:4}

\end{table}


\begin{table}[!t]
\centering
\caption{Black-box attack evaluation on Tiny-ImageNet dataset. The perturbation is generated on WideResNet-34-10.}
\resizebox{0.95\linewidth}{!}{
\begin{tabular}{cccccc}
\specialrule{.15em}{.1em}{.1em}
\multicolumn{2}{c}{\textbf{Method}}                         & \textbf{FGSM}  & \textbf{PGD-20} & \textbf{CW}    & \textbf{AutoAttack} \\ \hline
\multicolumn{1}{c}{\multirow{2}{*}{Madry}}  & Base & 45.42          & 46.09           & 47.36          & 46.27               \\
\multicolumn{1}{c}{}                                 & Ours & \textbf{47.45} & \textbf{47.99}  & \textbf{48.35} & \textbf{48.23}      \\ \hline
\multicolumn{1}{c}{\multirow{2}{*}{TRADESS}}  & Base & 46.30          & 47.23           & 48.15          & 48.52               \\
\multicolumn{1}{c}{}                                 & Ours & \textbf{48.45} & \textbf{49.03}  & \textbf{49.23} & \textbf{49.98}      \\ \hline
\multicolumn{1}{c}{\multirow{2}{*}{MART}}   & Base & 47.40          & 48.14           & 48.27          & 48.4               \\
\multicolumn{1}{c}{}                                 & Ours & \textbf{48.30} & \textbf{49.15}  & \textbf{49.61} & \textbf{49.50}      \\ \hline
\multicolumn{1}{c}{\multirow{2}{*}{HELP}} & Base & 48.91          & 49.47           & 49.88          & 50.01               \\
\multicolumn{1}{c}{}                                 & Ours & \textbf{49.71} & \textbf{49.93}  & \textbf{50.64} & \textbf{50.98}      \\
\specialrule{.15em}{.1em}{.1em}
\end{tabular}}

\label{tab:5}

\end{table}

\subsubsection{Black-box Evaluation}
Different from white-box adversarial attacks, black-box attacks generate adversarial perturbation from an unknown model. To show the robustness of the proposed proxy learning under the black-box attack settings, we pre-train WideResNet-34-10, then generate adversarial examples from the model. Then, we use the examples to our WideResNet-28-10 models. The black-box attack results are shown in Table \ref{tab:4} and \ref{tab:5}. `Base' denotes the reimplementation results with existing AT methods, and `Ours' denotes the results of proxy learning. Table \ref{tab:4} shows the black-box result on the CIFAR-10 dataset. As shown in the table, our method could improve the adversarial robustness of existing methods. Also, compared with the white-box results, we achieve better robustness and show close to the natural accuracy. Similar results are described in Table \ref{tab:5} conducted on the Tiny-ImageNet dataset.

\subsubsection{Sanity Check about Gradient Obfuscation}
Some works provide a false sense of security by vanishing the gradient. These methods are not considered to provide actual robustness. This phenomenon is called gradient obfuscation (gradient masking) \cite{athalye2018obfuscated}. It has been widely known that gradient obfuscation-based defense strategy does not provide adversarial robustness \mbox{\cite{athalye2018obfuscated, carlini2019evaluating, tramer2020adaptive, tramer2022detecting, kim2020revisiting}}. The adversarial robustness must guarantee their robustness under the worst case. However, the gradient obfuscation-based method supposes that the gradient of the model is not exposed and is unknown. Therefore, if the gradient of the model is exposed to the adversary, it cannot guarantee adversarial robustness. To handle this, it is important to verify whether the defense strategy is based on gradient obfuscation or not.
To verify that the defense strategy is not a gradient obfuscation, in \cite{athalye2018obfuscated, carlini2019evaluating}, they provide several sanity checks as follows:
\begin{itemize}
    \item It should show better robustness against 1-step attacks (\textit{e.g.,} FGSM) than iterative attacks (\textit{e.g.,} PGD).
    \item It should show better robustness against black-box attacks than white-box attacks.
\end{itemize}
To verify whether the defense strategy is based on gradient obfuscation or not, many studies have utilized the above points \mbox{\cite{tramer2020adaptive, tramer2022detecting,kim2020revisiting, zhou2022enhancing}}. Then, many defense strategies failed to pass the sanity check. However, from the above experiments, we verify that our proposed method does not suffer from gradient obfuscation. 1) In Table \ref{tab:3}, the proposed defense method shows better robustness against the FGSM attack than the PGD attack. 2) By comparing the result of Table \ref{tab:3}, Table \ref{tab:4}, and Table \ref{tab:5}, we verified that it shows better robustness under the black box setting than in the white-box setting. Through the sanity check, we demonstrate that the proposed method does not rely on gradient obfuscation.

\begin{table}[!t]
    \centering
    \caption{Robustness comparison results with recently proposed representation learning-based methods on ResNet-18 model with CIFAR-10.}
    \begin{tabular}{cccc}
\specialrule{.15em}{.1em}{.1em}
\textbf{Method}               & \textbf{FGSM}        & \textbf{PGD}         & \textbf{AutoAttack}          \\ \cmidrule{1-4}
TLA            & 60.1        & 50.5        & 47.0          \\ 
RoCL           & 60.5        & 51.6        & 48.8        \\
AdvCL          & 61.6          & 53.2          & 49.8        \\ 
AGKD-BML       & 61.3          & 52.3          & 50.0        \\ \cmidrule{1-4}

\textbf{Ours} & \textbf{62.5} & \textbf{55.3} & \textbf{51.2} \\ \specialrule{.15em}{.1em}{.1em}
\end{tabular}%
\label{tab:6}
\end{table}

\begin{table}[!t]
    \centering
    \caption{Comparison results with recently proposed methods that exploit the robust and non-robust features. The experiment was conducted on CIFAR-10 dataset with ResNet-18 model.}
    \begin{tabular}{ccccc}
\specialrule{.15em}{.1em}{.1em}
\textbf{Method}               & \textbf{FGSM}        & \textbf{PGD}    & \textbf{CW}      & \textbf{AutoAttack}          \\ \cmidrule{1-5}
CAFE          & 60.5          & 54.5          & 50.4        & 48.5        \\ 
DRRDN       & 62.1          & 52.1          & 51.3      &47.9        \\ \cmidrule{1-5}

\textbf{Ours} & \textbf{63.8} & \textbf{55.3} & \textbf{52.0} & \textbf{51.2}\\ \specialrule{.15em}{.1em}{.1em}
\end{tabular}%
\label{tab:minor_1}
\end{table}





\subsection{Comparison with Recently Proposed Methods} 
Many previous works try to enhance feature representation to improve adversarial robustness. In this section, we compare the proposed robust proxy learning method with recently proposed methods that enhance the feature representations (TLA \cite{mao2019metric}, RoCL \cite{kim2020adversarial}, AdvCL \cite{fan2021does}, AGKD-BML\cite{wang2021agkd}). In the case of TLA and AGKD-BML, they are fully supervised learning methods that apply existing metric learning frameworks to AT framework. In the case of RoCL and AdvCL, they apply AT framework to a self-supervised or unsupervised learning scheme for pretraining. Then, finetuned in a supervised manner. Table \ref{tab:6} shows the comparison results. As shown in the table, our proposed method achieves better robustness than other methods. Since our proposed method explicitly learns adversarially robust feature representation, it shows better robustness.

Furthermore, recently some works try to improve the robustness by exploiting the robust features \mbox{\cite{yang2021adversarial, Kim_2023_CVPR}}. Table \mbox{\ref{tab:minor_1}} shows the comparison results. DRRDN \mbox{\cite{yang2021adversarial}} denotes a method that disentangles the features into class-specific features and class-irrelevant features. Then, exploit class-specific features as robust features. In the case of CAFE \mbox{\cite{Kim_2023_CVPR}}, it extracts robust features by adversarial instrumental variable (IV) regression. The experiment was conducted on the CIFAR-10 dataset with the ResNet-18 model. As shown in the table, our proposed method shows better robustness than recently proposed methods that exploit robust features. 

\subsubsection{Fine-tune with Proxy Loss} As we referred above, many existing feature representation learning methods exploit their methods as pretraining in an unsupervised/semi-supervised manner. Then, finally, they fine-tune the model in a supervised manner. Therefore, the proposed method can be combined with existing representation learning as a fine-tuning process. To this end, we pretrain the model with AdvCL \cite{fan2021does} and RoCL \cite{kim2020adversarial}. Then, we fine-tuned the model with the proposed proxy loss. The results are shown in Table \ref{tab:7}. As shown in the table, our method can be combined with AdvCL and RoCL and improve the robustness over them. Furthermore, we verify the effectiveness of the proposed method under unseen adversarial attack ($L_2$-norm bounded attack). As shown in Table \ref{tab:7}, our proposed method still shows better robustness under unseen attack. The results can be interpreted that the proposed method can ensure robustness against unseen types of perturbations. Since we explicitly learned a feature that is resistant to noise variation, it shows robustness against unseen perturbations.

\begin{table}[t!]
\caption{Improving the robustness by combining the proposed method with existing representation learning-based methods}
\resizebox{0.98\linewidth}{!}{
\begin{tabular}{ccccccc}
\specialrule{.15em}{.1em}{.1em}
Pretraining            & Fine-tune     & FGSM          & PGD-$L_{\infty}$          & AutoAttack    & \begin{tabular}[c]{@{}c@{}}PGD-$L_{2}$\\ ($\epsilon=0.25$)\end{tabular} & \begin{tabular}[c]{@{}c@{}}PGD-$L_{2}$\\ ($\epsilon=0.5$)\end{tabular} \\ \hline
\multirow{2}{*}{RoCL}  & Base          & 60.5          & 51.6          & 48.8          & 70.3                                                     & 60.8                                                    \\
                       & \textbf{Ours} & \textbf{61.5} & \textbf{53.1} & \textbf{50.0} & \textbf{72.5}                                            & \textbf{62.9}                                           \\ \hline
\multirow{2}{*}{AdvCL} & Base          & 61.6          & 53.2          & 49.8          & 71.1                                                     & 60.0                                                    \\
                       & \textbf{Ours} & \textbf{62.8} & \textbf{54.8} & \textbf{51.1} & \textbf{73.2}                                            & \textbf{63.0}                                           \\ \specialrule{.15em}{.1em}{.1em}
\end{tabular}
}\label{tab:7}
\end{table}

\subsubsection{Computation Cost for Data Sampling in Proxy Learning}
Another advantage of Robust Proxy Learning compared to existing representation learning-based AT methods is that the complexity of training computation is low. Here, the training complexity represents the amount of computation required to solve the entire training dataset. Let $N$ denote the number of samples in the training dataset. Since most of existing methods take a tuple of data as a unit input (anchor, positive, negative), it requires high training complexity. For example, in the case of AdvCL and RoCL that exploit contrastive loss, they take a pair of data as input thus they require $O(N^2)$ training complexity. Furthermore, in the case of TLA that exploits Triplet loss, it takes triplets of data thus it requires $O(N^3)$ training complexity. Compared with these methods, in our method, we generate proxies for each class and compare every proxy with all samples. Therefore, the training complexity of the proposed method is $O(NC)$ where $C$ denotes the number of classes. Since $C\ll N$, the training complexity of the proposed method is much less than others.

\begin{table}[!t]
\centering
\caption{Robustness evaluation when the adversarial perturbation is generated by maximizing the gradient of non-robust channels}
\begin{tabular}{clc}
\specialrule{.15em}{.1em}{.1em}
Dataset                        & Method       & Accuracy      \\ \hline
\multirow{8}{*}{CIFAR-10}      & Madry        & 20.1          \\
                               & Madry+Proxy  & \textbf{23.2} \\ \cline{2-3} 
                               & TRADES       & 24.8          \\
                               & TRADES+Proxy & \textbf{27.1} \\ \cline{2-3} 
                               & MART         & 22.3          \\
                               & MART+Proxy   & \textbf{25.7} \\ \cline{2-3} 
                               & HELP         & 20.6          \\
                               & HELP+Proxy   & \textbf{23.8} \\ \hline
\multirow{8}{*}{Tiny-ImageNet} & Madry        & 9.2           \\
                               & Madry+Proxy  & \textbf{12.5} \\ \cline{2-3} 
                               & TRADES       & 11.9          \\
                               & TRADES+Proxy & \textbf{14.1} \\ \cline{2-3} 
                               & MART         & 12.7          \\
                               & MART+Proxy   & \textbf{14.1} \\ \cline{2-3} 
                               & HELP         & 11.2          \\
                               & HELP+Proxy   & \textbf{14.3} \\ \specialrule{.15em}{.1em}{.1em}
\end{tabular}
\label{tab:8}
\end{table}

\subsection{Attacking by Maximizing Gradient of Non-robust Features}
One way to directly attack the proposed method is to maximize the gradient of non-robust channels. To generate adversarial perturbation, we modify the Eq. \ref{eq:5} as follows:
\begin{equation}
\mathcal{L}_{attack}=-\mathcal{L}_{base}(f(x+p),y)-\left\|\mathcal{G}_{nr} \right\|_{2} + \left\|p \right\|_{2},
\label{eq:9}
\end{equation}
where $p$ is an adversarial perturbation that maximizes the gradient of non-robust channels and leads to misclassification. The robustness evaluation result is described in Table \ref{tab:8}. In the experiment, we set the perturbation budget as $\epsilon=0.03$. As shown in the table, when maximizing the gradient of non-robust channels, the baseline models (Madry, TRADES, MART, and HELP)  are significantly broken. However, when we train the model with the proposed proxy learning (Madry+Proxy, TRADES+Proxy, etc), it shows better robustness. Since the feature that reduces the effect of the non-robust channels is explicitly learned, it is not easily attacked even if we maximize them, and it shows better robustness.

\begin{figure*}[!t]
	\centering
	    \includegraphics[width=0.75\linewidth]{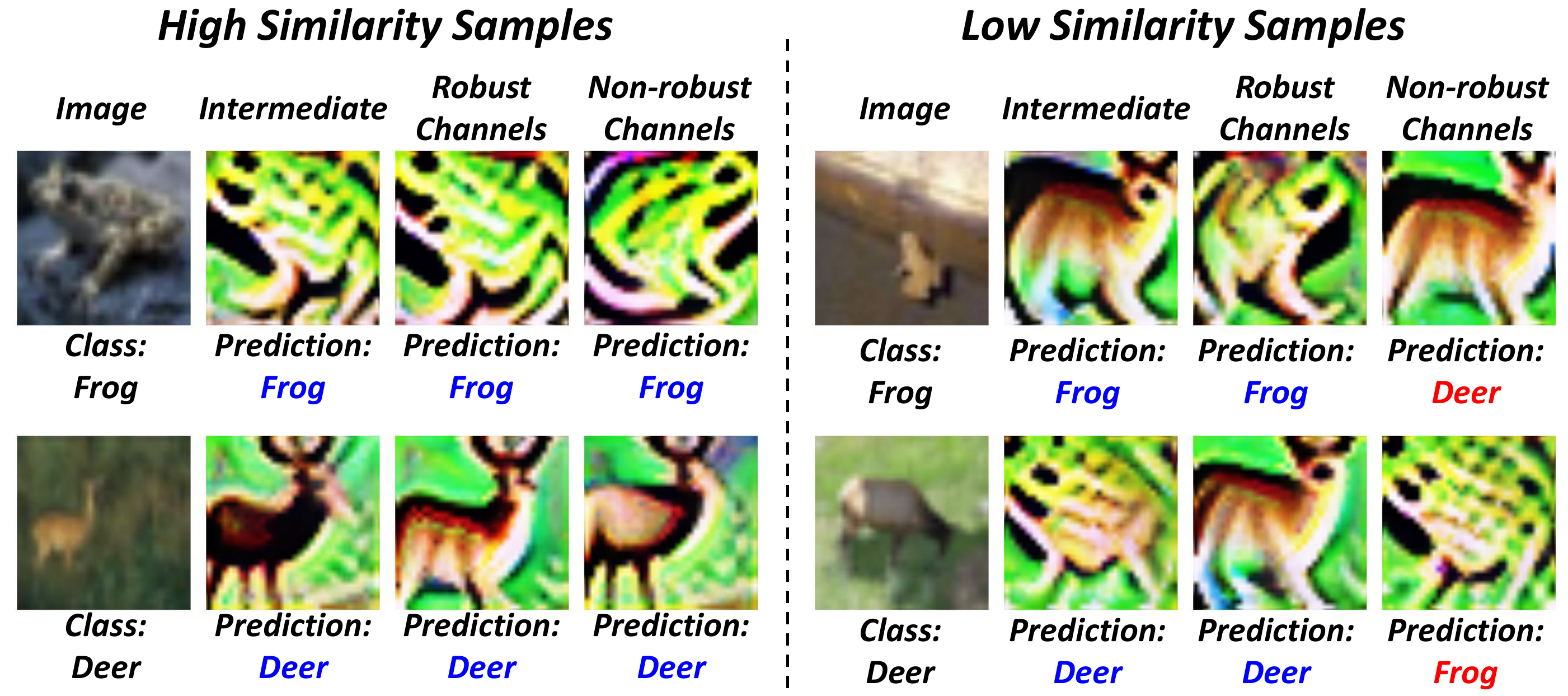}
	\caption{Visualization of learned representation. The `Intermediate' denotes the visualization results of the intermediate feature, and the `Robust Channels' denotes the visualization results of a set of channels with rarely be manipulated by adversarial perturbations. The `Non-robust Channels' denotes the visualization results of the set of channels that can potentially be manipulated by adversarial perturbation.} 
	\label{fig:7}
\end{figure*}

\subsection{Effectiveness of Robust Proxy}
In this section, we verify the effectiveness of robust proxy by ablation study. To this end, we conduct an ablation study by using different types of proxies in Eq.\mbox{\ref{eq:prox}}. Table \mbox{\ref{tab:r2_2}} shows the experimental results with the ResNet18 model on CIFAR-10 and Tiny-ImageNet datasets when use different types of proxies. In the table, Madry+\mbox{$\textrm{Proxy}_{\textrm{normal}}$} denotes the results when using a normal instance as the anchor, Madry+\mbox{$\textrm{Proxy}_{\textrm{avg}}$} denotes the experimental results when using proxy by averaging all robust features. Also, Madry+\mbox{$\textrm{Proxy}_{\textrm{ours}}$} denotes the results when using our proposed robust proxies as the anchor. As shown in the table, when using the robust proxy, it shows better robustness than when using normal instances as the anchor. Since the normal instances do not guarantee robust representation, they cannot improve the adversarial robustness. However, in the proposed method, the robust proxies have robust feature representation, we could improve the robustness by robust proxy learning. Furthermore, when using the average of robust proxies, it shows less robustness compared to using the proposed robust proxy. Since simply averaging the features could lead to different feature representations, it could lead the robustness decrement.

\begin{table}[!t]
\centering
\caption{Ablation study by using different types of proxies. The experiment was conducted on the ResNet18 model with CIFAR-10 and Tiny-ImageNet.}
\begin{tabular}{clcccc}
\specialrule{.15em}{.1em}{.1em}
\multicolumn{1}{c}{Dataset}    & \multicolumn{1}{c}{Method} & FGSM          & PGD           & CW            & AA            \\ \hline
\multirow{3}{*}{CIFAR-10}      & Madry + $\textrm{Proxy}_{\textrm{normal}}$              & 60.1          & 50.5          & 48.1          & 47.0          \\
                               & Madry + $\textrm{Proxy}_{\textrm{avg}}$              & 61.0          & 51.7          & 48.5          & 47.1          \\
                               & Madry + $\textrm{Proxy}_{\textrm{ours}}$              & \textbf{62.0} & \textbf{53.1} & \textbf{49.4} & \textbf{48.9} \\ \hline
\multirow{3}{*}{Tiny-ImageNet} & Madry + $\textrm{Proxy}_{\textrm{normal}}$              & 23.5          & 21.1          & 18.0          & 17.0          \\
                               & Madry + $\textrm{Proxy}_{\textrm{avg}}$              & 23.5          & 20.2          & 19.0          & 18.0          \\
                               & Madry + $\textrm{Proxy}_{\textrm{ours}}$              & \textbf{24.1} & \textbf{22.7} & \textbf{20.5} & \textbf{19.9} \\ \specialrule{.15em}{.1em}{.1em}
\end{tabular}
\label{tab:r2_2}
\end{table}

\subsection{Visual Interpretation of Learned Representations}
\subsubsection{Feature Visualization} In this section, we show the advantage of the proposed method from the lens of learned feature representation. To this end, we visualize the features that have high similarity and low similarity to proxies. We visualize the features by \cite{kim2021distilling}. In \mbox{\cite{kim2021distilling}}, the visualization results are optimized from random image. It optimizes the random input image so that the features extracted from the optimized input image are similar to the robust or non-robust features. Then, the optimized image becomes the visualization result. Fig. \ref{fig:7} shows the visualization results of robust and non-robust channels for each input. In the figure, following the definition of \cite{kim2021distilling}, the `Intermediate' denotes the visualization results of the intermediate feature, and the `Robust Channels' denotes the visualization results of a set of channels with rarely be manipulated by adversarial perturbations. `Non-robust Channels' denotes the visualization results of the set of channels that can potentially be manipulated by adversarial perturbation. As shown in the figure, the features of `High Similarity Samples' have semantic information by themselves, and the features of `Low Similarity Samples' do not have semantic information by themselves. Specifically, in the case of non-robust channels of high similarity samples, it remains the semantical information of ground-truth classes. Therefore, the prediction does not change even if adversarial perturbations are added. In contrast to, in the case of the negative units of low similarity samples, it contains different semantical information from the ground-truth class. Therefore, it is easily manipulated by adversarial perturbation.
\begin{figure}[!t]
	\centering
	    \includegraphics[width=0.85\linewidth]{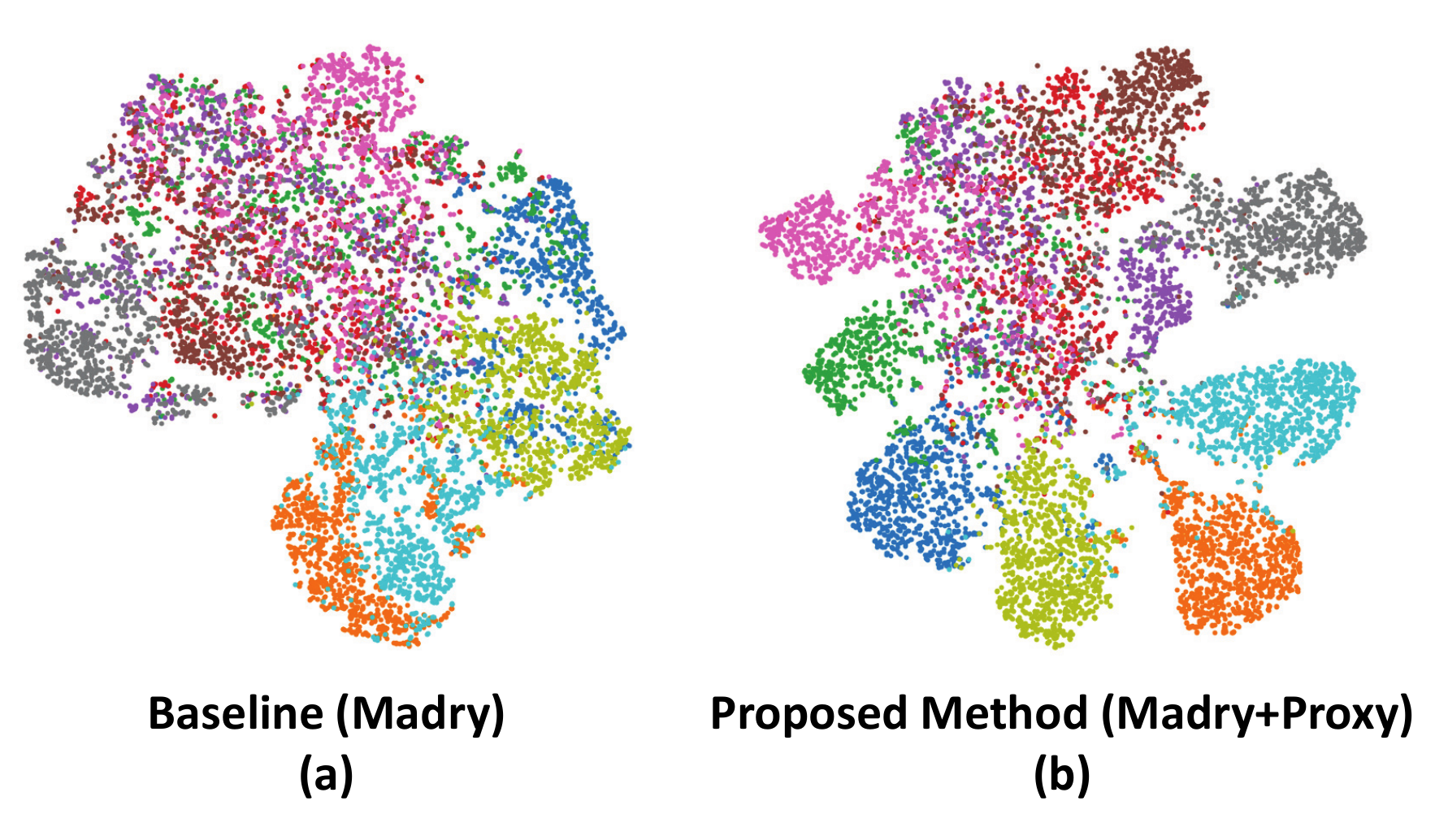}
	\caption{t-SNE visualization of representations according to the differently trained model. (a) denotes the t-SNE feature visualization results of the existing method (Madry). (b) denotes the t-SNE feature visualization results of the proposed method (Madry+Proxy).} 
	\label{fig:10}
\end{figure}
\begin{figure}[!t]
	\centering
	    \includegraphics[width=0.98\linewidth]{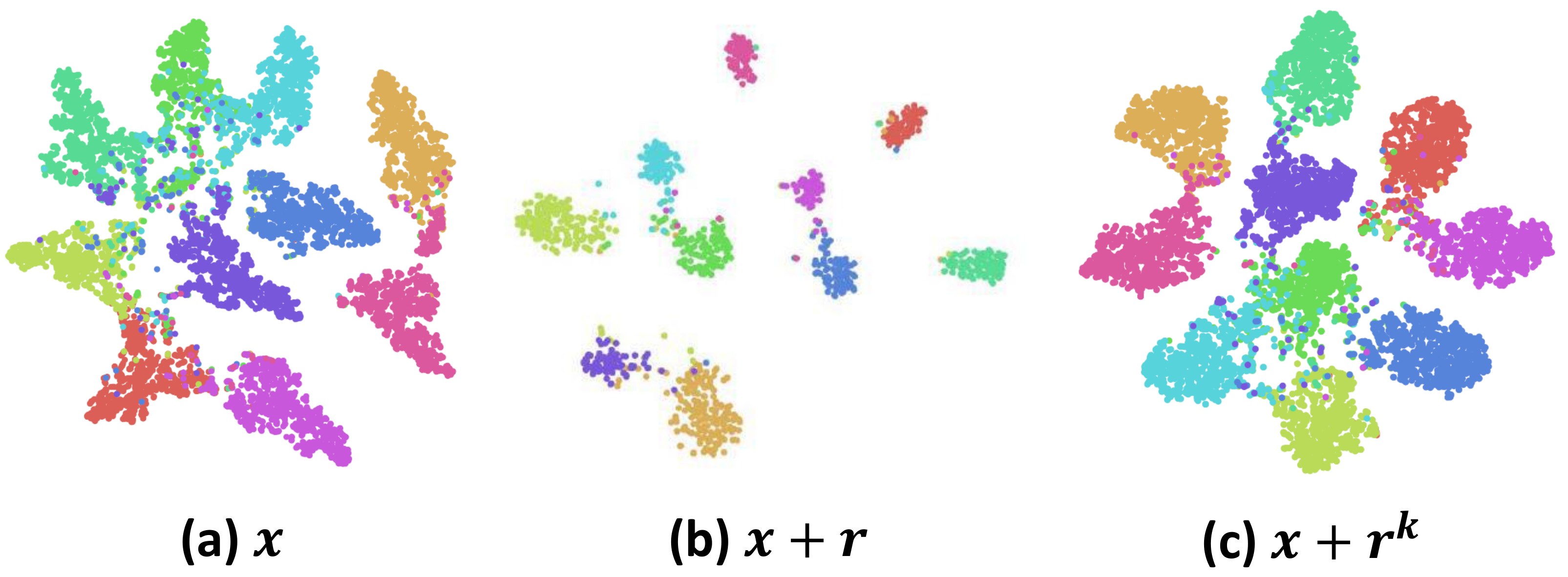}
	\caption{t-SNE visualization of representations according to different input types. Adding $r$ or $r^k$ gives a much clearer separation among classes.} 
	\label{fig:8}
\end{figure}

\subsubsection{Verify the effectiveness of Proposed Proxy Learning by t-SNE visualization}
In the proposed method, the performance improvement mainly stems from the separation of robust and non-robust features. Different from previous works, since our proposed method explicitly learns class discriminative robust features, it shows better robustness. To verify this, we visualize the features using t-SNE on the CIFAR-10. Fig. \mbox{\ref{fig:10}} shows the t-SNE results. Fig. \mbox{\ref{fig:10}}  (a) shows features extracted from the base model (Madry). Fig. \mbox{\ref{fig:10}} (b) shows features extracted from the model trained by our proposed method. As shown in the figure, Fig. \mbox{\ref{fig:10}} (b) shows more discriminative feature distribution and clearer class boundary than (a). This can be interpreted that the proposed method can learn more robust and class discriminative features.
\subsubsection{Verify the effectiveness of CRP by t-SNE visualization}
To further demonstrate the efficacy of CRP, we visualize the features according to the input types using t-SNE on the CIFAR-10 dataset. Fig. \ref{fig:8} shows the t-SNE results. As shown in the figure, when we add $r$ or $r^k$ to input, it shows a much clearer class boundary than using the original images. This can be interpreted that $r$ and $r^k$ make the adversary difficult to successfully attack an image, leading to more robust prediction.

\begin{figure}[!t]
	\centering
	    \includegraphics[width=0.85\linewidth]{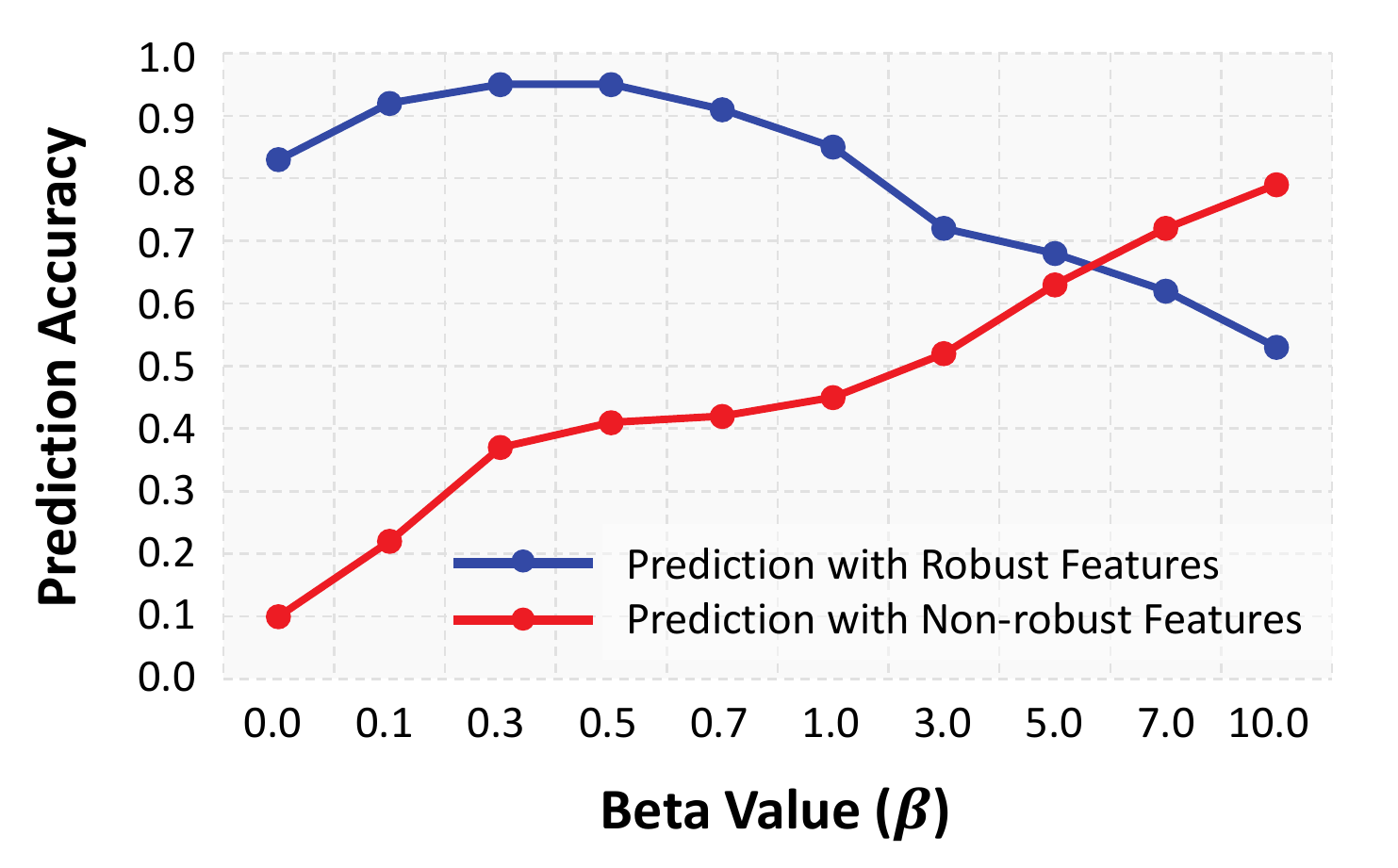}
	\caption{The accuracy comparison of robust and non-robust features by controlling information flows of features. The blue line denotes the accuracy when predicting only with robust features and the red line denotes the accuracy when predicting only with non-robust features. Note that $\beta$ regulates the total amount of the information that flows into the features.}
	\label{fig:9}
\end{figure}
\subsection{Effect of Information Bottleneck in Feature Distillation}
In this section, we conduct ablation studies to verify the role of the information bottleneck. To this end, we change the $\beta$ values in Eq. \mbox{\ref{eq:1}} to analyze how the information bottleneck controls the information flow of the robust and non-robust features. In the equation, $\beta$ regulates the total amount of the information that flows into the features. Increasing the $\beta$ value means that minimizing the information flow of the robust features while maximizing the information flow of the non-robust features. Therefore, we will compare classification accuracy for the robust and non-robust features according to the $\beta$ value. Fig. \mbox{\ref{fig:9}} shows the results when predicting with distilled robust or non-robust features. In the figure, the blue line represents the accuracy of predictions using only the robust features. Also, the red line represents the accuracy of predictions using only the non-robust features. As shown in the figure, as the $\beta$ value increases, the accuracy of predicting with the robust feature decreases. Since the amount of information flowing into the robust feature decreases as the $\beta$ value increases, the prediction accuracy with robust features decreases. On the other hand, as the $\beta$ value increases, the accuracy of predicting with non-robust features increases since the amount of information flowing into the non-robust feature increase. Through the experiment, we demonstrated how the information bottleneck affects the information flow of the robust and non-robust features.


\section{Discussion}
In this research, we generate robust proxies that have robust feature representations. Then, we train the model so that the features resemble the representation of those proxies. For future direction, there are many ways that could improve the proposed method. For example, it is possible to train a network that only leverages robust features by manually masking the non-robust features during the training. Also, exploiting multiple proxies from the same class or generating a more reliable proxy could be a good future direction.

\section{Conclusion}
 In this paper, we introduce the intriguing, yet not explored aspect of adversarial training that explicitly learns adversarially robust features. Many works have demonstrated that adversarial vulnerability mainly stems from the non-robust components of learned features, while how to explicitly learn robust features is not explored. To tackle the problem, we manually generate adversarially robust features and propose a novel training framework called robust proxy learning that explicitly learns robust features. To this end, through the CEO algorithm, we generate class representative robust features called robust proxies.  During the training, DNNs explicitly learn the representation of robust proxies through the proposed robust proxy learning framework. For each proxy, we pull the data of the same class close to the proxy and push others away in the feature space, allowing the model to explicitly learn adversarially robust features. Extensive experimental results suggest that the proposed method can improve the robustness of existing AT methods under stronger attacks and be general and flexible enough to be adopted on any AT
methods. We believe that the proposed method could shed new insight into utilizing robust perturbation for adversarial robustness.

\bibliographystyle{IEEEtran}
\bibliography{egbib}


\vfill

\begin{IEEEbiography}[{\includegraphics[width=1in,height=1.25in,clip]{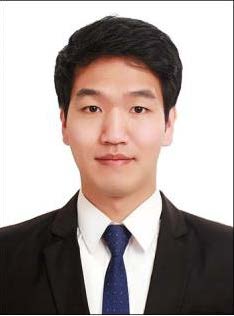}}]{HONG JOO LEE}
received the B.S. degree from Ajou University, Suwon, South Korea, in 2016, and the M.S. and Ph.D. degrees from the Korea Advanced Institute of Science and Technology (KAIST), Daejeon, South Korea, in 2018 and 2023. His research interests include deep learning, machine learning, medical image segmentation, and adversarial robustness.
\end{IEEEbiography}

\begin{IEEEbiography}[{\includegraphics[width=1in,height=1.25in,clip]{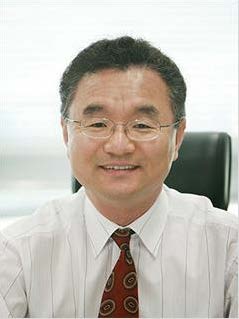}}]{YONG MAN RO}
(Senior Member, IEEE) is a Professor of Electrical Engineering and the Director of the Center for Applied Research in Artificial Intelligence (CARAI) at the Korea Advanced Institute of Science and Technology (KAIST). He received his B.S. degree from Yonsei University, Seoul, South Korea, and his M.S. and Ph.D. degrees from KAIST. He has been a Researcher at Columbia University, a Visiting Researcher at the University of California at Irvine, and a Research Fellow of the University of California at Berkeley. He was also a Visiting Professor with the Department of Electrical and Computer Engineering, University of Toronto, Canada. His research interests span image and video systems, including image processing, computer vision, multimodal learning, vision-language learning, and object detection. He received the Young Investigator Finalist Award of ISMRM in 1992 and the Year’s Scientist Award (Korea) in 2003. He is an Associate Editor for IEEE TRANSACTIONS ON CIRCUITS AND SYSTEMS FOR VIDEO TECHNOLOGY and has served as an Associate Editor for IEEE SIGNAL PROCESSING LETTERS. He has also served as a TPC member, program chair, and special session organizer for many international conferences.
\end{IEEEbiography}

\end{document}